\documentclass[journal]{IEEEtran} 

\usepackage{amsmath,amsfonts}
\usepackage{algorithm}
\usepackage{array}
\usepackage[caption=false,font=normalsize,labelfont=sf,textfont=sf]{subfig}
\usepackage{textcomp}
\usepackage{stfloats}
\usepackage{url}
\usepackage{verbatim}
\usepackage{graphicx}
\usepackage{cite}
\usepackage{paralist}
\usepackage{comment}
\usepackage{todonotes}
\usepackage{hyperref}
\usepackage{microtype} 
\usepackage{booktabs}
\usepackage{siunitx}
\usepackage{changepage}
\usepackage{multirow}
\usepackage{bigdelim}
\usepackage{MnSymbol}
\usepackage{balance}

\usepackage{algpseudocode}

\begin{document}

\title{PRO-MIND$^{\star}$: Proximity and Reactivity Optimisation of robot Motion to tune safety limits, human stress, and productivity in INDustrial settings}

\author{Marta Lagomarsino,
Marta Lorenzini,
Elena De Momi, and
Arash Ajoudani 
\thanks{$^{\star}$ The term PRO-MIND recalls our aim to promote the mental well-being of human workers and, at the same time, denotes the advanced intelligence of the robotic partner.}
\thanks{Manuscript received: Month, DD, YYYY; Revised Month, DD, YYYY; Accepted Month, DD, YYYY.}%
\thanks{This paper was recommended for publication by Editor Name Surname upon evaluation of the Associate Editor and Reviewers' comments.}%
\thanks{This work was supported in part by the ERC-StG Ergo-Lean (Grant Agreement No.850932), in part by the European Union’s Horizon 2020 research and innovation programme SOPHIA (Grant Agreement No. 871237). The authors thank Dr. Fabio Fusaro for his help in the implementation.} 
\thanks{Marta Lagomarsino is with the Human-Robot Interfaces and Interaction Laboratory, Istituto Italiano di Tecnologia, Genoa, Italy and Department of Electronics, Information and Bioengineering, Politecnico di Milano, Milan, Italy. Corresponding author's email: {\tt\small marta.lagomarsino@iit.it}}%
\thanks{Marta Lorenzini and Arash Ajoudani are with the Human-Robot Interfaces and Interaction Laboratory, Istituto Italiano di Tecnologia, Genoa, Italy.}%
\thanks{Elena De Momi is with the Department of Electronics, Information and Bioengineering, Politecnico di Milano, Milan, Italy.}%
\thanks{Digital Object Identifier (DOI): NA.}
\vspace{-0.5cm}
}

\markboth{Journal of \LaTeX\ Class Files,~Vol.~XX, No.~XX, Month~2024}%
{Lagomarsino \MakeLowercase{\textit{et al.}}: A Sample Article Using IEEEtran.cls for IEEE Journals}

\maketitle

\begin{abstract}
Despite impressive advancements of industrial collaborative robots, their potential remains largely untapped due to the difficulty in balancing human safety and comfort with fast production constraints. To help address this challenge, we present PRO-MIND$^{\star}$, a novel human-in-the-loop framework that leverages valuable data about the human co-worker to optimise robot trajectories. By estimating human attention and mental effort, our method dynamically adjusts safety zones and enables on-the-fly alterations of the robot path to enhance human comfort and optimal stopping conditions. Moreover, we formulate a multi-objective optimisation to adapt the robot's trajectory execution time and smoothness based on the current human psycho-physical stress, estimated from heart rate variability and frantic movements. These adaptations exploit the properties of B-spline curves to preserve continuity and smoothness, which are crucial factors in improving motion predictability and comfort. Evaluation in two realistic case studies showcases the framework's ability to restrain the operators' workload and stress and to ensure their safety while enhancing human-robot productivity. Further strengths of PRO-MIND include its adaptability to each individual's specific needs and sensitivity to variations in attention, mental effort, and stress during task execution. 
\end{abstract}

\begin{IEEEkeywords}
Human Factors and Human-in-the-Loop; 
Human-Centered Robotics; 
Cognitive Human-Robot Interaction; 
Motion and Path Planning
\end{IEEEkeywords}

\section{Introduction}
\label{sec:introduction}

\IEEEPARstart{T}{he} advent of Industry 5.0 allows humans and machines harmoniously collaborate to foster resource-efficient and user-centred manufacturing. This paradigm surpasses mere yield and productivity, elevating the well-being of workers to the forefront of industrial processes \cite{Maddikunta2021}. To materialise this ambitious vision, collaborative robots (CoBots) have gained considerable traction across industries. 
CoBots could alleviate the physical burden placed on human counterparts \cite{sirintuna2022carrying} or provide support in hazardous operations \cite{liu2020remote}. 
However, still today, the potential of CoBots is largely underexploited due to the difficulty of integrating human safety and perceived comfort requirements with classical production constraints, such as fast cycle time capabilities. 

Safety in human-robot collaborative applications is currently regulated by internationally recognised standards, such as ISO 10218 \cite{iso10218-1, iso10218-2}, and more recently, the technical specification ISO TS 15066 \cite{iso15066}, which introduced two criteria: Speed and Separation Monitoring (SSM) and Power and Force Limiting (PFL).
The SSM standard restricts the robot speed based on the relative distance and velocity between humans and robots, often resulting in delays and non-essential idle times when the CoBot approaches the human worker. 
Conversely, the PFL configures thresholds, e.g. on forces, torques, velocities, and mechanical power, for reducing risks associated with transient contact. 
Yet, since it does not consider the separation distance, it can reduce performance when the operator is far away from the robot or when the robot moves away from the human. 
The aforementioned limitations may impact the efficiency and performance of the system.

In addition to ensuring safety, a primary objective in hybrid industrial scenarios is to prevent CoBots from negatively affecting the psycho-physical state of their human counterparts \cite{lorenzini2023ergonomic, Gualtieri2021}. This can be accomplished by designing legible, predictable, and smooth CoBot movements, which may resemble human gestures, alongside carefully planning and managing the space sharing and interaction between them. 
Indeed, many activities do not inherently pose physical safety hazards for the workers (such as the risk of direct harm by the robot), but create an environment conducive to the development of cumulative work-related stress, discomfort, and time pressures. Currently, these disorders affect hundreds of millions of people worldwide \cite{who2022}, compromising operators' performance and leading to direct financial implications for both private companies and governments \cite{Hassard2014}. 
But, the prospect of CoBots goes beyond the peaceful coexistence with humans \cite{lorenzini2023ergonomic}. 
CoBots may help reduce the mental demands of workers by relieving them from repetitive and monotonous tasks, contribute to error prevention by actively detecting signs of tiredness and inattention in human operators, and regulate the production rhythm in accordance with the worker's capabilities and alertness.  

While prioritising safety and comfort is essential, an overly conservative approach in human-robot collaborative applications can inadvertently lead to human frustration and unnecessarily limit workplace productivity. Indeed, restrictions on force and speed, along with the emphasis on smoothness and legibility, may result in slow and wide movements even when they are not needed.
By employing monitoring of human attention and psycho-physical state through non-invasive and practical sensor systems, this paper proposes adjusting the proximity level and reactive behaviour of the CoBot to promote efficient collaboration while enhancing workers' well-being, both at the physical level (e.g. avoiding collisions) and the socio-cognitive level (e.g. fostering cognitive-grounded safety\footnote{Throughout this paper, the term ``physical safety'' management pertains to preventing unintended physical contact between humans and robots. On the other hand, ``cognitive-grounded safety'' describes the human subjective perception of the safety level during the interaction, as proposed in \cite{kirschner2022expectable}.} and rediscovering an optimal psycho-physical state).

A preliminary study in this direction was proposed in \cite{Lagomarsino2022robot}, where the adaptation of CoBot velocity in response to changes in Heart Rate Variability (HRV) was introduced as a strategy to tackle the stress-productivity trade-off. However, this earlier version of our method overlooked some crucial factors: 
\begin{inparaenum}[(i)]
    \item potential safety concerns, raised by directly transmitting desired joint positions and velocities to the CoBot without accounting for the current position of the human co-worker;
    \item the human perception of safety, disregarding that the comfortable separation distance perceived by each subject may change with the awareness of the CoBot motion and current mental effort \cite{kim2021construction}; 
    \item limited stress detection to physiological measures, leading to delays in identifying stressful conditions.
\end{inparaenum} \\
In contrast, the present study surpasses these limitations by online adapting the path and timing law of CoBot trajectories based on the multi-factor estimation of human awareness and psycho-physical state. 
Using the local modification and strong convex hull properties of B-splines, we smoothly adjust the nominal path and ensure compliance with both physical and cognitive-grounded safety zones around the human. 
In addition, we integrated the theory of camera-based monitoring of human attention level and body movements into our framework, providing deeper insights into the operator's experience and anticipating stress detection. 

Overall, the main contributions of this study can be summarised as follows.
\begin{enumerate}
    \item We proposed an adaptive system that can respond to individual differences in cognitive resources and stress responses, generating personalised CoBot trajectories that comply with the task advancement.
    \item We exploited camera-extracted data regarding the human co-worker, such as their attention to the CoBot and the estimated mental effort, to adapt physical and cognitive-grounded safety zones online while ensuring the fluency of the collaboration.
    \item We designed a multi-objective optimisation problem and an ECG and camera-based decision-making strategy to optimise the total execution time and smoothness of the CoBot trajectory to maximise productivity without perilously increasing workers’ psycho-physical stress.
    \item The architecture was tested with multi-subject experiments in two realistic case studies (manual collaborative tasks) performed with a $7$ degrees of freedom manipulator.
    \item We compared the method with state-of-the-art approaches and demonstrated the practical benefit of the proposed method by considering fluency and productivity metrics, as well as well-known biomarkers of psycho-physical stress and subjective questionnaires.  
\end{enumerate}

The rest of the paper is structured as follows. 
In \autoref{sec:state-of-the-art}, we reviewed the literature on designing adaptive CoBot behaviours to promote human safety and comfort. 
Next, we summarised the main contributions and presented the salient features of PRO-MIND (\autoref{sec:contribution}). Details about the developed trajectory planner and the strategy to alter on-the-fly the nominal path according to human attention and mental effort are described in \autoref{sec:trajectories} and \ref{sec:path_adaptation}, respectively. This is followed in \autoref{sec:timing_law_adaptation} by defining a multi-objective optimisation problem to find the optimal trade-off between human psycho-physical stress and workplace productivity.
Experiments are then proposed in  \autoref{sec:experiments}, and results are validated through statistical analysis (\autoref{sec:result}).
Final sections (\ref{sec:discussion} and \ref{sec:conclusion}) discuss the framework limitations and future research directions.

\section{Related Works}
\label{sec:state-of-the-art}

\subsection{Ensure Human Safety through Robot Adaptation}

Trajectory generation and safety management are longstanding problems, thoroughly investigated in the literature.
Safety constraints have been tackled by modifying the generated trajectory using estimates of the robot stopping time \cite{scalera2020application}, employing safety barrier functions \cite{landi2019safety}, or designing dodging motions for the robot \cite{zanchettin2019towards}. 
Recent studies propose adaptively switching safety zones around the robot, which are adjusted based on its speed  \cite{karagiannis2022adaptive, scalera2021optimal} or motion direction \cite{scalera2022enhancing}. 
The work in \cite{safeea2019online} models the robot and human co-worker as capsules and alters on-the-fly the robot nominal path to prevent unintended contact with the human. With the same aim, alternative execution paths are designed in \cite{zanchettin2019towards} by inserting waypoints along the motion and in \cite{Scoccia2021acollision} interpolating Bézier curves.

Traditionally, planning minimum-time trajectories on desired paths for higher throughput and scaling the robot velocity profile for safety handling have been addressed separately, resulting in suboptimal solutions.
The challenge lies in determining the optimal strategy for deciding when the robot must inevitably slow down or even stop to ensure safety while preventing productivity reduction unless necessary, especially in scenarios where the human and the robot work side-by-side.
An iterative procedure based on human position and speed is proposed in \cite{Palleschi2021} to safely replan the time evolution of the robot motion along the assigned path. 
In \cite{zanchettin2022safe}, the velocity scaling approach leverages the dynamic parameters of the robot, rather than just lumped characteristics as stopping times or distances. \\
Both works emphasise the need for a unified approach that handles productivity (minimum-time execution) and safety simultaneously. However, these studies overlook human perception of safety, individual attitudes and familiarity with the technology. This may prevent user acceptance and undermine the potential for the successful integration of robots into the workplace. 

In \cite{lasota2015analyzing, beschi2020how}, the impact of motion planning parameters on human perception has been investigated. \cite{lasota2015analyzing} demonstrated that when robot motion planning is aware of the next human task, it is possible to effectively reduce idle times and enhance user satisfaction.
Additionally, \cite{beschi2020how} found that operators prefer to decide the task timing. Interestingly, the dependability of robot motion (i.e. its trustworthiness and perceived reliability) is not determined by the velocity profile but rather by the cycle time. Consequently, in \cite{faroni2022safety}, the same authors introduced a cost function to minimise the expected path execution time, accounting for safety standard-based speed reduction due to human proximity, but without further examination of human perception. 
Moreover, there is limited exploration of valuable online data regarding the human co-worker, such as their current awareness of the robot motion, which could provide significant insights for planning a more seamless and efficient collaboration. 
\\
In a first attempt to establish the cognitive-grounded safety concept, \cite{kirschner2022expectable} analysed human Involuntary Motion Occurrences (IMO) and developed a model that relates human expectation to the separation distance and robot velocity. This was used to limit velocity such that the IMO did not exceed an acceptable threshold. 
Yet, online knowledge about the human state was not exploited, and the models lacked personalisation.

\subsection{Enhance Human Psycho-physical State through Robot Adaptation}

\begin{figure*}[!b]
    \includegraphics[width=\linewidth]{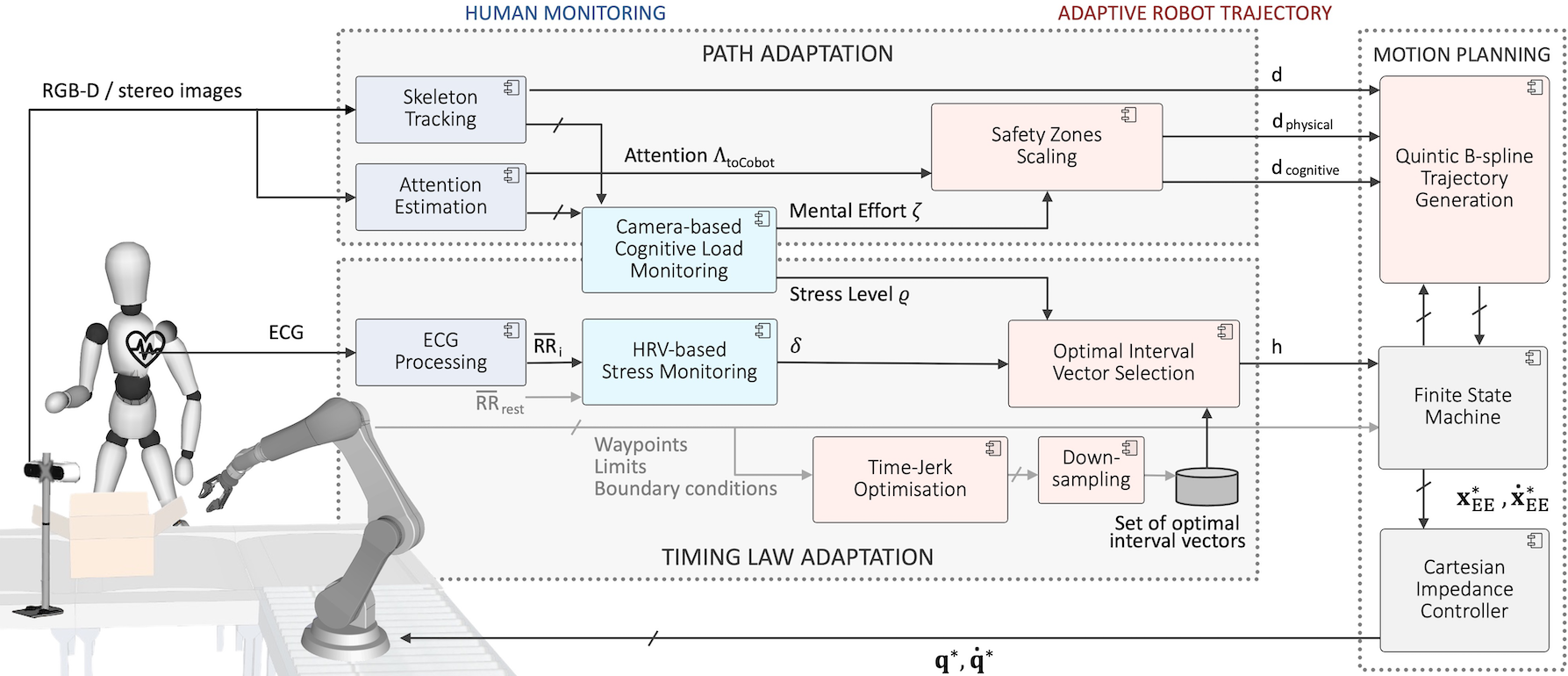}
    \caption{PRO-MIND framework adapting CoBot trajectory to promote human well-being and system productivity. Note that computational blue modules refer to human monitoring, red modules determine the path and timing adaptation of the CoBot trajectory, and variables in grey are provided offline.}
    \label{fig:schema}
\end{figure*}

A large and growing body of literature has focused on modelling human mental processes \cite{lu2022mental, lorenzini2023ergonomic} and quantifying task execution costs \cite{Haji2015, hu2020interact, hu2022toward}. 
The measurements generally fall into three categories: physiological measures, subjective rating scale, and performance-based measures. Nevertheless, these analyses are typically conducted offline to evaluate the overall quality of the interaction retrospectively. 

To enhance workers' well-being, it is crucial to design HRC systems capable of online assessing the mental effort and psycho-physical stress of the operator and adapting the behaviour of the robotic teammate accordingly. 
Research following this principle, usually referred to as ``affective robotics'' \cite{spitale2023affective, braezal2016social} (i.e. the combination of robotics and affective computing), relies on the monitoring and interpretation of nonverbal communication, such as attention\cite{Lagomarsino2022anonline, Zhang2023thehuman}, body language\cite{Lagomarsino2022pick, wang2022computational} and physiological signals (i.e. cardiac \cite{ayres2021thevalidity, landi2018relieving} and electrodermal \cite{Hopko2021} activity). Continuous information gathering about the operator's mental processes is necessary due to the subjective and multidimensional nature of psycho-physical states. Additionally, as mental fatigue accumulates, the operator's needs may change throughout task execution. 

A pioneering example exploiting the human affect estimate for an industrial robotics application was found in \cite{landi2018relieving, Villani2018}, where the remote control of a mobile robot was simplified when the user's stress, estimated by analysing HRV, exceeded human capabilities. Specifically, the robot provided assistance to stressed users by reducing its maximum velocity. 
Similarly, \cite{shah2023robot} exploited a reinforcement learning problem in a close-proximity teleoperation task to enable assistance commands and alert signals. In this case, the prediction of cognitive fatigue was quantified by the confidence of a binary classification support vector machine (SVM) model, which used HRV features extracted from the electrocardiography (ECG) signal. 

To understand how humans prioritise ergonomics and conjoint efficiency in human-robot collaboration, \cite{pantano2023effects} examined users' expressed preferences among four available task configuration options. Their findings indicate that robotic experts prioritise teamwork over ergonomics, while novices act to minimise individual efforts and robot travel paths. 
In our recent work \cite{lagomarsino2023maximising}, we developed an adaptive system to learn on-the-fly interaction parameters' values that aligned with user preferences and maximise the conjoint efficiency. To this aim, the proposed approach considers human physical and cognitive factors and robot internal expenses into a unified reinforcement-learning framework to proactively optimise the robot actions online. 
\\
In \cite{Messeri2021}, HRC was treated as a repeated non-cooperative game, where the human and the robot were considered self-interested players with different objectives: stress minimisation for the human and productivity maximisation for the robot. The collaboration state was estimated, and the pace of the robot was adjusted accordingly. 

Despite the growing interest, few task-specific countermeasures have been proposed in contrast to numerous studies underlining the impact of robot behaviour on human well-being and performance \cite{Hopko2021}.  
There is still considerable room for improvement in this area, focusing not only on robot timing but also on adapting its travelled path and personalising assistance according to individual differences to enhance the overall collaboration experience. 

\section{
PRO-MIND Framework
}
\label{sec:contribution}
The proposed framework is outlined in \autoref{fig:schema}. Cartesian trajectories are generated through quintic B-spline curves, which feature continuous velocity, acceleration, and jerk along the path. 
Indeed, smooth robot manoeuvres have shown to be more legible \cite{dragan2013legibility} and contribute to a safer and more comfortable experience for human partners \cite{Kuehnlenz2016}, possibly attributed to their similarity to natural arm movements \cite{Flash1985}. On this line, the CoBot trajectories are adjusted on the fly exploiting B-spline properties to preserve the desired kinematic characteristics. The resulting adaptations are gradual and continuous, preventing sudden and unpredictable CoBot movements and optimising stopping conditions.

Regarding the \emph{Path Adaptation} block, we adaptively scale physical and cognitive-grounded\footnotemark[1] safety zones around the operator based on their level of awareness. 
More specifically, we detect the operator's attention towards the CoBot and estimate their mental effort directly from RGB images using our system validated in \cite{Lagomarsino2022anonline, Lagomarsino2022pick}. 
Based on this information, we enable online path alterations to prevent the CoBot from entering the cognitive-grounded safety zone, enhancing the perception of safety \cite{Arai2010}, and trigger a stop command if it violates the physical safety limits, preventing unintended contact between humans and CoBots.
In particular, the algorithm responds to distraction (or concentration on other tasks) and growth in mental demand with an increase in the level of safety, i.e. ensuring a larger human-robot separation distance or halting the motion.

Concerning the \emph{Timing Law Adaptation} block, a multi-objective optimisation problem is implemented to tune the duration (minimum-time) and smoothness (minimum-jerk) of the trajectories accomplished by the CoBot based on the human psycho-physical state. 
While minimising jerk can ensure cognitive ergonomics \cite{Rojas2020}, pure jerk minimisation may lead to slow robot motions, compromising the overall productivity of the human-robot dyad. Hence, we incorporate an objective to minimise the time required to follow the path. An online decision-making algorithm then selects the most appropriate solution, balancing a trade-off between CoBot execution speed and the user's psycho-physical stress. The latter is continuously monitored by analysing HRV and detecting frantic movements through the camera.

\section{Quintic B-spline Trajectories} 
\label{sec:trajectories}

The trajectory planning problem exploits the formulation of B-splines in the Cartesian space. 
A B-spline curve\footnote{For further details about B-spline curves, the reader is referred to \cite{Gasparetto2007}.} of degree $p$ is defined as:
\vspace{-0.2cm}
\begin{equation}
\label{eq:bspline}
    \textbf{x}_\text{EE}(t) = 
    \sum_{k=1}^{C+1} N_{k,p}(t) \, \textbf{c}_k
\vspace{-0.1cm}
\end{equation}
where $N_{k,p}(t)$ are basis functions expressed recursively by the De Boor formula \cite{Boor2001} through the definition of a sequence of $M+1$ knots $[\tau_1, \tau_2, \dots \tau_{M+1}]$ in the interval $t \in [0,t_\text{f}]$:
\begin{equation}
\label{eq:deboor}
\begin{cases}
    N_{k,p}(t) = \frac{t-\tau_k}{\tau_{k+p}-\tau_k} N_{k,p-1}(t) + \frac{\tau_{k+p+1}-t}{\tau_{k+p+1}-\tau_{k+1}} N_{k+1,p-1}(t) \\
    N_{k,0}(t) = 
    \begin{cases}
        1, & \text{if } \tau_k\leq t <\tau_{k+1} \\
        0, & \text{otherwise.}
    \end{cases}
\end{cases}
\end{equation} 
The linear combination in \autoref{eq:bspline} is defined by $C+1$ control points $\textbf{c}_k$, defining a polygon in which the curve $\textbf{b}(t)$ is contained (strong convex hull property). More specifically, if $t$ is in knot span $[\tau_l,\tau_{l+1})$, then $\textbf{b}(t)$ is in the convex hull of control points $\textbf{c}_{l-p}, \textbf{c}_{l-p+1}, \dots, \textbf{c}_l$. 

B-splines of degree $p$ present continuous derivatives up to the $\{p-1\}$-th derivative, i.e. $\mathcal{C}^{p-1}$.
Within the proposed control strategy, we leveraged B-spline curves to generate desired paths in the Cartesian space for the manipulator. Its end-effector pose $\textbf{x}_{\text{ee}}$ is defined by position $x$, $y$, and $z$ and orientation expressed through Euler angles, i.e. roll $\phi$, pitch $\vartheta$, yaw $\psi$. 
Formulating B-splines of degree five, we obtain end-effector trajectories with continuous forth-derivative of the Cartesian position and orientation, thus guaranteeing bounded values of velocity, acceleration and jerk along the path. 
Moreover, $p=5$ permits imposing boundary conditions till the acceleration. 

Another key feature of B-spline formalism is that changing the position of control point $\textbf{c}_l$ only affects the curve $\textbf{b}(t)$ in the knot interval $[\tau_l,\tau_{l+p+1})$. The local modification scheme is pivotal to curve design because it allows modifying the curve locally without changing its shape in a global way. 
Within this work (see \autoref{sec:path_adaptation}), we exploit this property to alter on-the-fly the nominal path of the manipulation to ensure the most appropriate distance to the human while ensuring continuity in the path and the associated kinematic quantities.

\begin{figure}[b!]
    \vspace{-0.4cm}
    \centering
    \includegraphics[width=\linewidth]{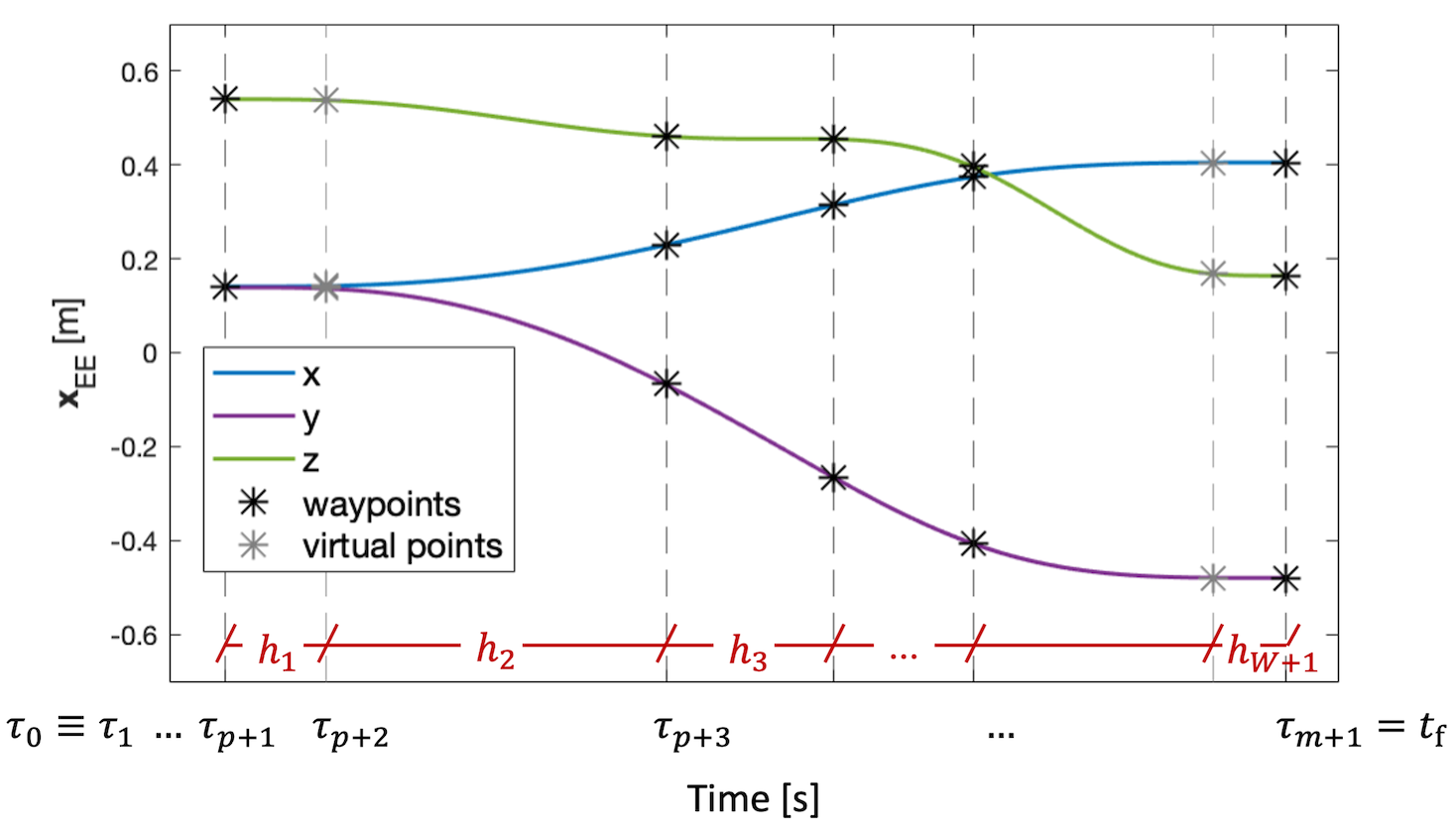}
    \vspace{-0.7cm}
    \caption{$x$, $y$, and $z$ coordinates of end-effector trajectory defined by a B-spline curve passing through a set of waypoints (black dots) and two virtual points (grey dots). Time intervals $h_l$ between consecutive points, defining optimisation vector $\bf{h}$, are highlighted.}
    \label{fig:bspline_image}
\end{figure}

Besides, a waypoint-based trajectory planner was designed to enhance the flexibility of the framework. This allows the user to quickly teach the CoBot a new task by manually positioning and orienting the end-effector in a sequence of desired poses called waypoints. 
As a result, each trajectory is subject to the condition of passing through $W$ waypoints $\{\textbf{w}_1, \textbf{w}_2, \dots, \textbf{w}_W\}^j$ at a sequence of unknown time instants $\{t_1, t_2,\dots, t_W\}$. 
In \autoref{sec:timing_law_adaptation}, the procedure to find the productivity-stress optimal time intervals between consecutive waypoints is defined. 

Given the intervals vector, the knot vector $\tau$ can be filled. The values at the extremities feature multiplicity $p+1$, so that the first and last control points coincide with the desired waypoints $\textbf{w}_1$ and $\textbf{w}_W$ and are attained at $t_1=0$ and $t_W=t_\text{f}$. 
Two virtual points have been additionally introduced to impose zero boundary conditions also for the jerk in fifth-degree curves. %
Consequently, $\textbf{x}_{\text{EE}}(t)$ is defined by a six-dimension B-spline with $p=5$ (e.g. $\textbf{x}_\text{EE}(t)=\textbf{b}(t) \in \mathbb{R}^6$), $\tau$ is composed by $M\!+\!1=(W\!+\!2)\!+\!2p=W\!+\!12\,$ knots (see \autoref{fig:bspline_image}) and the number of required control points is $C\!+\!1=W\!+\!p\!+\!1=W\!+\!6$. 

The derivative of a B-spline curve of degree $p$ is a B-spline of degree $p-1$. Therefore, the velocity, acceleration and jerk of the end-effector trajectory result in: 
{\allowdisplaybreaks
\begin{align}
    \dot{\textbf{x}}_\text{EE}(t) = 
        & \sum_{k=1}^{C} N_{k,p-1}(t) \, \textbf{c}_{k,1} 
        \label{eq:velocity} \\
    \ddot{\textbf{x}}_\text{EE}(t) = 
        & \sum_{k=1}^{C-1} N_{k,p-2}(t) \, \textbf{c}_{k,2} \\
    \dddot{\textbf{x}}_\text{EE}(t) = 
        & \sum_{k=1}^{C-2} N_{k,p-3}(t) \, \textbf{c}_{k,3}
\end{align}
}
where the control points of $d$-th derivative are expressed as: 
\begin{equation}
    \textbf{c}_{k,d} = 
    \begin{cases}
        \textbf{c}_{k} & \text{if } d=0\\
        (p+1-d)\dfrac{\textbf{c}_{k,d-1}-\textbf{c}_{k-1,d-1}}{\tau_{k+p+1-d}-\tau_k} & \text{otherwise.}
    \end{cases}
\end{equation}

\section{Path Adaptation}
\label{sec:path_adaptation}

This section proposed a path adaptation strategy to ensure physical and cognitive-grounded safety according to human attention and estimated mental effort. 

\subsection{Online Camera-based Human Monitoring}
\label{sec:camera}

The human position during the interaction and their level of awareness are computed directly from the input images of a low-cost RGB-D or stereo camera. 
The choice of the external sensor system was motivated by the desire to avoid wearability constraints. 
Noteworthy, the module is scalable to other person tracking methods, such as inertial-based motion-capture systems\cite{lagomarsino2023maximising}.

\begin{figure*}[!t]
    \centering
    \includegraphics[width=0.62\linewidth]{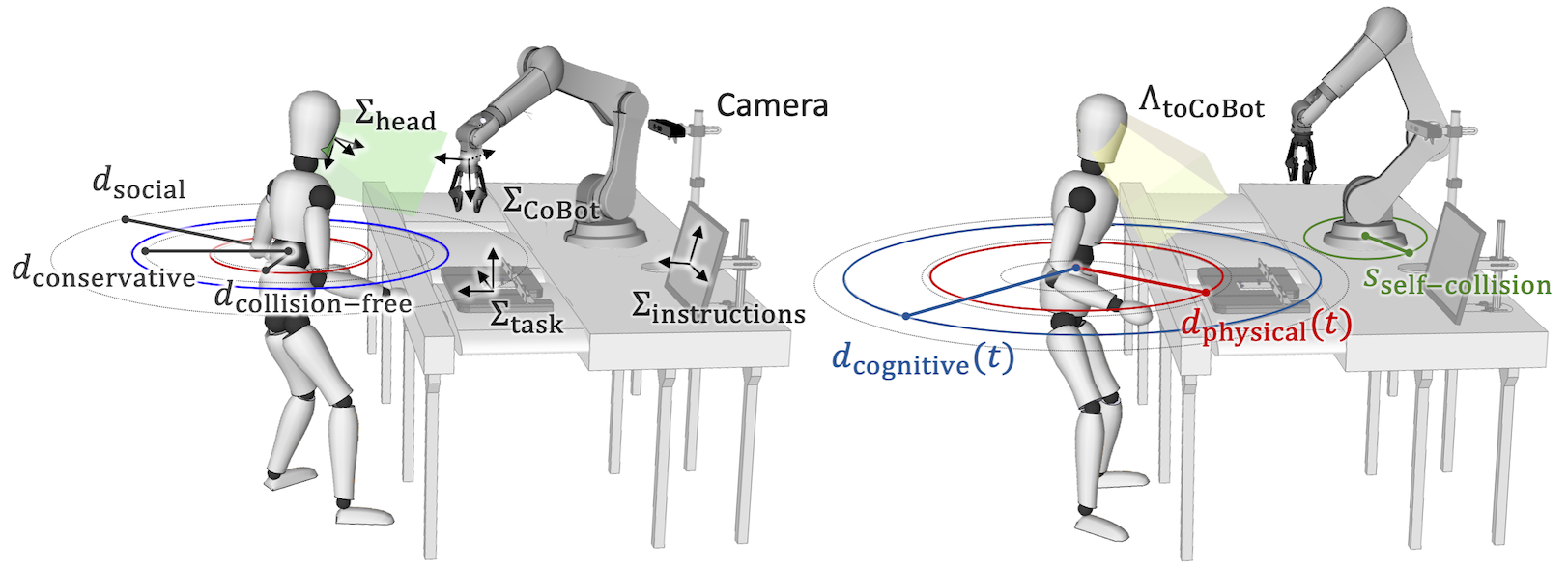}
    \hspace{0.1cm}
    \includegraphics[width=0.36\linewidth]{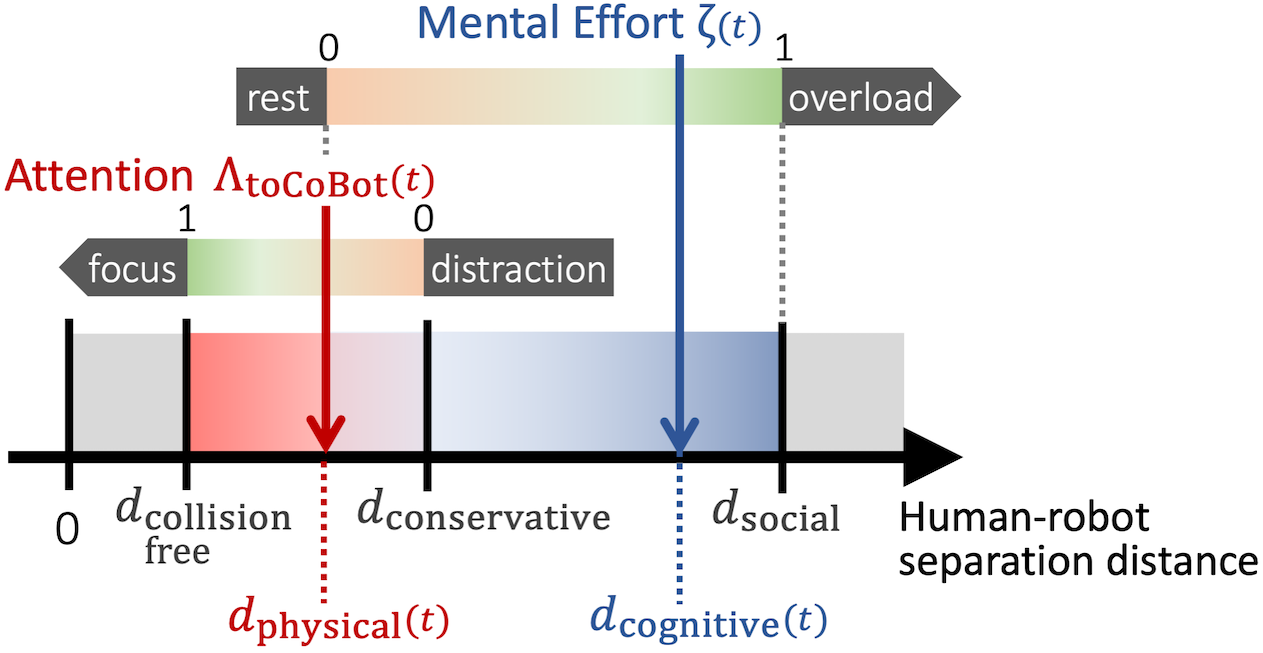}
    \vspace{-0.6cm}
    \caption{Conceptual illustration exemplifying the dynamic scaling of physical and cognitive-grounded safety zones based on human awareness.}
\label{fig:safety_zones2}   
\end{figure*}

A visual 3D skeleton tracking algorithm developed by StereoLabs was adopted to track the human position over time and measure the separation distance to the CoBot. 
The RGB images captured by the camera are also employed to detect the human face and identify facial landmarks, using the OpenCV library and a TensorFlow pre-trained deep learning model, respectively. Consequently, the head pose is continuously estimated by solving a Perspective-n-Point (PnP) problem between the OpenFace 3D model of the face and the detector output and stabilising the pose using a Kalman Filter. 
Thus, a frame $\Sigma_\text{head}$ is associated with the user's head (see \autoref{fig:safety_zones2}).

To estimate the attention level towards a given Region of Interest (RoI) to which a frame $\Sigma_\text{RoI}$ is associated, we map the Cartesian vector expressing the relative position between $\Sigma_\text{head}$ and $\Sigma_\text{RoI}$ into spherical coordinates (azimuth angle $\theta_\text{RoI}(t)$, elevation angle $\varphi_\text{RoI}(t)$, and radial distance $r_\text{RoI}(t)$). 
A fuzzy logic function exploiting the Raised-Cosine Filter \cite{Lagomarsino2022anonline} is then applied to separately normalise the measured attention angles: 
\begin{equation}
\label{eq:attention}
\!\! f(\alpha{\scriptstyle (t)})\! =\!
\begin{cases}
    1, 
        & {\displaystyle \!\!\!\! 
            \mbox{if } \! \left\lvert\alpha{\scriptstyle (t)}\right\rvert \leq\alpha^{\text{min}}{\scriptstyle (t)}} \\
        
    \frac{1}{2} \Bigl[1 \!
        - \cos{\Bigl(\dfrac{\left\lvert\alpha{\scriptstyle (t)}\right\rvert - \alpha^{\text{min}}{\scriptstyle (t)}}
        {\alpha^{\text{max}}{\scriptstyle (t)} - \alpha^{\text{min}}{\scriptstyle (t)}}
        \pi\Bigr)} \Bigr], 
        & {\displaystyle \!\!\!\!\! \underset{\displaystyle 
            \& \left\lvert\alpha{\scriptstyle (t)}\right\rvert \leq \alpha^{\text{max}}{\scriptstyle (t)}}
            {\mbox{if } \!   \left\lvert\alpha{\scriptstyle (t)}\right\rvert\! 
            >\alpha^{\text{min}}{\scriptstyle (t)}
            }} \\
    0, 
        & {\displaystyle \!\!\!\!
        \mbox{otherwise.}}
\end{cases}
\end{equation}
Note that $\alpha$ indicates $\theta_\text{RoI}$ or $\varphi_\text{RoI}$ at each time instant $t$, and the threshold values $\theta_\text{RoI}^{\text{min}}{(t)}$, 
$\theta_\text{RoI}^{\text{max}}{(t)}$,
$\varphi_\text{RoI}^{\text{min}}{(t)}$,
$\varphi_\text{RoI}^{\text{max}}{(t)}$ depend on the RoI size and the current distance from the human operator. 
Specifically, the lower and upper limits are computed as:
\begin{equation}
\label{eq:attention_limits}
    \alpha^{\text{min}}{(t)}=\text{tan}^{-1}\!\left(\!\frac{(1\!-\!\gamma) \,a}{2 \, r_\text{RoI}{\scriptstyle (t)}}\!\right)\!, 
    \,
    \alpha^{\text{max}}{(t)}=\text{tan}^{-1}\!\left(\!\frac{(1\!+\!\gamma) \,a}{2 \, r_\text{RoI}{\scriptstyle (t)}}\!\right)\!, 
\end{equation}
where $a$ represents the width (if $\alpha\!=\!\theta_\text{RoI}$) or height (if $\alpha\!=\!\varphi_\text{RoI}$) of the RoI and $r_\text{RoI}(t)$ denotes the distance between $\Sigma_\text{head}$ and $\Sigma_\text{RoI}$. 
The function $f(\alpha{\scriptstyle (t)})$ has a smooth behaviour (from $1$ to $0$) in the angle range corresponding to the tangential distances spanning in $[ (1\!-\!\gamma)\, a, \; (1\!+\!\gamma)\, a]$ with respect to $\Sigma_\text{RoI}$ and $\gamma \in \mathbb{R}$. 

The attention level $\Lambda_\text{RoI}(t)$ towards the RoI is determined by the product of the normalised azimuth and elevation indicators:
\vspace{-0.2cm}
\begin{equation}
    \Lambda_\text{RoI}(t) = \Lambda(\theta_\text{RoI}{\scriptstyle (t)}, \varphi_\text{RoI}{\scriptstyle (t)})\, =  f(\theta_\text{RoI}{\scriptstyle (t)}) \, f(\varphi_\text{RoI}{\scriptstyle (t)}).
\end{equation} 

This formulation allows us to monitor whether the human is currently distracted or concentrated on a particular RoI. It is worth noting that $\Sigma_\text{RoI}$ can be linked to both static and dynamic attention areas, which can vary in shape and size. 
In this study, we assess the human's awareness of the CoBot motion by monitoring their attention towards the CoBot end-effector (i.e. a dynamic attention area), denoted as $\Lambda_\text{toCoBot}$.
Moreover, we exploit our cognitive load assessment framework proposed and validated in \cite{Lagomarsino2022anonline, Lagomarsino2022pick} to assess the \textit{mental effort} during the interaction. 
With this aim, we define two additional static attention areas within the workplace, i.e. $\Sigma_\text{task}$ defining the manual task area, and $\Sigma_\text{instructions}$ providing information and step-by-step instructions to accomplish the task, typically through a monitor. 
Hence, we examine the following factors: 
\begin{itemize}
    \item \textit{Wariness for Assistant}: This metric quantifies the number of glances and gazes toward the CoBot motion over time, namely the transitions in human attention between $\Sigma_\text{CoBot}$ and other focal points over time:
    \vspace{-0.2cm}
    \begin{equation}
    \small
        m_\text{WA}(t) = \sum_{i=1}^{N_\text{checks}} \frac{\text{[instant of CoBot check]}_i}{\text{time elapsed}}.
    \label{eq:ME1}
    \end{equation}
    \item \textit{Learning Delay}: Inspired by Rapid Instructed Task Learning theory \cite{Cole2012}, this factor evaluates the capacity for rapid learning of new activity and the operator's automaticity in task completion. A prolonged focus on the activity indicates slower learning and higher mental demand:
    \vspace{-0.2cm}
    \begin{equation}
    \small
        m_\text{LD}(t) = \frac{\text{dwell time on }\Sigma_\text{task}}{\text{time elapsed}}. 
    \label{eq:ME2}
    \end{equation}
    \item \textit{Instruction Cost}: This factor examines the instructions' quality to gather information about the work. It involves counting the switches in attention between $\Sigma_\text{task}$ and $\Sigma_\text{instructions}$, except those needed to check new instructions:
    \vspace{-0.2cm}
    \begin{equation}
    \label{eq:ME3}
    \small
        m_\text{IC}(t) = \sum_{i=1}^{N_\text{switches}} \dfrac{\text{[instant of not required switch]}_i}{\text{time elapsed}}.
    \end{equation}
\end{itemize}
The first factor reflects the human's lack of confidence and trust in the robotic partner by relying on research on gaze tracking and interpretation. This mistrust during the collaboration has been found in the literature to usurp mental resources \cite{Crawford2019advances}.
The remaining factors provide insights into task advancement and the mental demand experienced by the human. Primary focus on the manual task area or frequent checks of the instructions indicates incomplete accomplishment of the manual activity. To accommodate this valuable information in our path adaptation strategy, we introduce a comprehensive score called \textit{mental effort} $\zeta$. The latter combines the normalised and weighted factors (\autoref{eq:ME1},\ref{eq:ME2},\ref{eq:ME3}) using thresholds and coefficients obtained from our previous study \cite{Lagomarsino2022anonline}. 
\vspace{-0.4cm}

\subsection{Safety Zones Scaling}

In this section, we propose an online strategy to adapt distance thresholds and define physical and cognitive-grounded safety zones around the human based on their level of awareness.
The idea behind this is derived from prior research findings indicating that the preferred distance between humans and robots is subjective and strongly influenced by the individual's current condition.
Drawing inspiration from proxemics theory \cite{hall1966hidden}, we define three distance thresholds: $d_\text{collision-free}$, which represents the minimum distance to avoid hazards; $d_\text{conservative}$, ensuring physical safety more cautiously (defining the boundary between \textit{intimate} and \textit{personal} zones presented in \cite{hall1966hidden}); and $d_\text{social}$, marking the beginning of the \textit{social} zone (see \autoref{fig:safety_zones2}). The radius of the physical safety zone around the human, denoted as $d_\text{physical}(t)$, spans from $d_\text{collision-free}$ to $d_\text{conservative}$, while the radius of the cognitive-based safety zone $d_\text{cognitive}(t)$ ranges from $d_\text{physical}(t)$ to $d_\text{social}$.

Specifically, we define $d_\text{physical}(t)$ to be proportional to the percentage of attention $\Lambda_\text{toCoBot} \in [0,1]$ that the human directs towards the CoBot:
\vspace{-0.3cm}
\begin{equation}
\label{eq:d_physical}
 d_\text{physical}{\scriptstyle (t)} = 
 d_\text{conservative} - 
 \Lambda_\text{toCoBot}{\scriptstyle (t)}
 \bigl(d_\text{conservative} - d_\text{collision-free}\bigr) .
\end{equation}
In practical terms, when the human is focused on the CoBot, the physical safety zone limit decreases, allowing for closer proximity, up to $d_\text{physical} = d_\text{collision-free}$ when $\Lambda_\text{toCoBot}\!=\!1$. Conversely, the algorithm responds to distraction or concentration on other tasks (i.e. decrease in $\Lambda_\text{toCoBot}$ up to 0) ensuring a larger separation distance between the human and the CoBot, with $d_\text{physical}$ reaching a maximum value of $d_{\text{conservative}}$. 

Furthermore, to enhance human comfort during the interaction, we employ the \textit{mental effort} score $\zeta \in [0,1]$ introduced in the previous subsection.
This permits defining the appropriate interaction distance that the CoBot should maintain to ensure human cognitive-grounded safety:
\begin{equation}
\label{eq:d_cognitive}
 d_\text{cognitive}{\scriptstyle (t)} = \!\max \!
    \Big[
    d_\text{physical}{\scriptstyle (t)}, 
    \bigl(d_\text{social} - d_\text{conservative}\bigr) \zeta{\scriptstyle (t)} + d_\text{conservative}
    \Big].
\end{equation}
\vspace{-1cm}

\subsection{Fine Curve Shape Control}

\begin{figure}[!t]
    \centering
    \includegraphics[width=\linewidth]{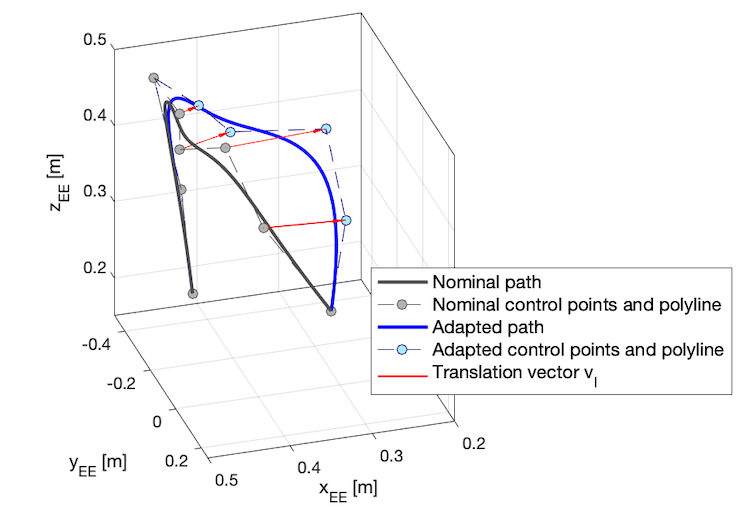}
    \vspace{-0.5cm}
    \caption{3D-view of end-effector trajectory highlighting control points and their modification while the trajectory is accomplished.}
    \label{fig:bspline3d}
\end{figure}

Our objective is to dynamically alter the path to maintain the separation distance $d_\text{cognitive}(t)$ that caters to current human awareness and mental effort, and to potentially stop whenever $d_\text{physical}(t)$ is suddenly violated. However, it is crucial to ensure that these adjustments do not compromise the continuity and smoothness of the CoBot path. We acknowledge the well-established influence of path smoothness on human perception, and we are committed to preserving this aspect during the interaction. Therefore, we leverage the local modification and strong convex hull properties of B-spline curves to design and edit the shape of the CoBot path based on the current human position and \textit{mental effort}. 
As mentioned earlier, modifying the position of control point $\textbf{c}_l$ only affects the curve $\textbf{b}(t)$ within the interval $[\tau_l,\tau_{l+p+1})$. 
So, at time $t$ in $[\tau_\text{ix}, \tau_\text{ix+1})$, we modify the control points $\textbf{c}_\text{ix+1}, \textbf{c}_\text{ix+2}, \dots, \textbf{c}_{W-p+8}$ to attain the desired $d_\text{cognitive}$ while preserving path continuity. Moreover, by leaving the last $p-1$ control points unchanged, we ensure that the trajectory reaches the final waypoint $\textbf{w}_W$, which remains unaltered, respecting the boundary conditions. 
The modification is performed as follows:
\vspace{-0.1cm}
\begin{align}
    \Bar{\textbf{c}}_{l}(t) 
    =  & \,
    \textbf{c}_{l}(t) + \textbf{v}_l(t) 
    =  \, \textbf{c}_{l}(t) + \bigl(d_\text{cognitive}(t) - d_l(t)\bigr) \, \hat{\textbf{v}}_l(t), \nonumber \\
    & l \in [\text{ix}+1, W\!-\!p\!+\!8)
\label{eq:modification-control-points}
\end{align}
where $d_l$ and $\hat{\textbf{v}}_l$ denote the module and the direction respectively of the vector between the human and the $l$-th control point on the horizontal plane (i.e. $\hat{\textbf{v}}_l$ is the unit vector).
The shape change caused by this modification is translational, occurring in the direction of the moved control point (see \autoref{fig:bspline3d}). 
Specifically, when the control point $\textbf{c}_l$ is moved to the new position $\bar{\textbf{c}}_l$ by adding the vector $\textbf{v}_l$, the original curve $\textbf{b}(t)$ is deflected in the same direction where $N_{l,p}(t)$ is non-zero, namely on the interval $[\tau_l,\tau_{l+p+1})$: 
\begin{align}
    \Bar{\textbf{b}}(t) 
    & = \sum_{k=1}^{l-1} N_{k,p}(t) \, \textbf{c}_k 
    + N_{l,p}(t) \, \bigl(\textbf{c}_l + \textbf{v}_l\bigr) 
    + \sum_{k=l+1}^{C+1} N_{k,p}(t) \, \textbf{c}_k \nonumber \\
    & = \sum_{k=1}^{C+1} N_{k,p}(t) \, \textbf{c}_k 
    + N_{l,p}(t) \, \textbf{v}_l \nonumber \\
    & = \textbf{b}(t) + N_{l,p}(t) \, \textbf{v}_l.
\end{align}
Although the distance moved varies from point to point, the strong convex hull property guarantees precise control over the shape of the curve. 
To ensure that the alteration of control points does not lead to self-collisions, we perform a check using a threshold $s_\text{self-collision}$ around the CoBot. If any violation is detected, we adjust the modification in \autoref{eq:modification-control-points} as follows: 
\begin{align}
    \Bar{\textbf{c}}_{l}(t) 
    = 
    \Bar{\textbf{c}}_{l}(t) + \bigl(s_\text{self-collision} - s_l(t)\bigr) \, \hat{\textbf{u}}_l(t)
\end{align}
where $s_l$ and $\hat{\textbf{u}}_l(t)$ represent the distance and direction between the CoBot base and end-effector on the horizontal plane. 

If the path adaptation is not implemented promptly or the human suddenly approaches the CoBot, violating the physical safety threshold distance $d_{\text{physical safety}}$, a warning procedure is initiated for the CoBot. We first examine the relative direction on the horizontal plane between the CoBot end-effector velocity vector and the position vector connecting the human and the CoBot end-effector. The CoBot will continue its path if the angle between these vectors exceeds a threshold $\beta_\text{th}$, indicating a significant divergence. On the other hand, if the angle is below $\beta_\text{th}$, revealing potential proximity, a CoBot stop command is triggered. Once the safety condition has been reestablished, we generate a B-spline trajectory that originates from the current position of the end-effector and passes through the remaining untraveled waypoints.

\section{Timing Law Adaptation}
\label{sec:timing_law_adaptation}

This section introduces an adaptive timing law that addresses the trade-off between the psycho-physical stress induced on the human operator and system productivity. 
By monitoring the user's stress level during the interaction, the interaction pace and the smoothness of CoBot trajectories can be adjusted and optimised to enhance human-robot collaboration. 

\subsection{Multi-objective Time/Jerk Trajectory Optimisation}
\label{sec:optimisation}

The CoBot trajectory, defined by a B-spline curve in the Cartesian space, is optimised by simultaneously minimising the total execution time and the integral of the squared jerk (i.e. maximising the smoothness) along the trajectory. 
It is crucial to notice that the two objectives have conflicting effects: reducing the former results in rapidly completed trajectories that feature large kinematic quantities values, while lowering the latter leads to smoother trajectories taking more time to reach the target pose. 
Our goal is to select the most appropriate solution based on the current human psycho-physical condition. The problem is formulated as a multi-objective optimisation task:
\begin{equation}
\begin{aligned}
    \underset{\textbf{h}}{\text{min}}  \quad & f_\text{time}(\textbf{h})=D\,\sum_{l=1}^{W+1}{h_l} \\
    \underset{\textbf{h}}{\text{min}}  \quad & f_\text{jerk}(\textbf{h})=\sum_{j=1}^{D}{\int_{0}^{t_\text{f}} \left(\sum_{k=1}^{C-2}  N_{k,p-3}(t) \, \textbf{c}_{k,3}\right)^2 dt}
    \\
    s.t. \quad & |\dot{\textbf{x}}_\text{EE}(t)|\leq \textbf{v}^{\text{max}} \\
    \quad & |\ddot{\textbf{x}}_\text{EE}(t)|\leq \textbf{a}^{\text{max}} \\
    \quad & |\!\dddot{\textbf{x}}_\text{EE}(t)|\leq \textbf{j}^{\text{max}}\\
    \quad & h_g\geq h_g^{\text{lb}} \ \quad\quad\quad\quad g=1,\dots,W+1 \\
\end{aligned}
\end{equation}
where the variables 
that should be optimised  are the time intervals between consecutive waypoints (see \autoref{fig:bspline_image}), 
denoted as $h_g=\tau_{p+g+1}-\tau_{p+g}$, 
and are grouped in the decision vector: 
\vspace{-0.1cm}
\begin{equation}
    \textbf{h}=[h_1, h_2,\dots, h_{W+1}].
    \vspace{-0.1cm}
\end{equation}
Note that the integral of the squared jerk in $f_{\text{jerk}}$ can be expressed in analytical form as a function of \textbf{h} by reporting the basis function $N_{k,p-3}$ through the recursive formula in \autoref{eq:deboor}. However, the computation is extremely verbose, thus a numerical integration procedure can also be exploited at the cost of increasing the computation time.

The optimisation problem is subject to inequality constraints on the Cartesian velocity, acceleration, and jerk. These constraints can be expressed in terms of the corresponding control points, $\textbf{c}_{k,1}$, $\textbf{c}_{k,2}$,  $\textbf{c}_{k,3}$, by exploiting the convex hull property of B-spline curves:
\begin{equation}
\begin{aligned}
    |\textbf{c}_{k,1}|= & 
    \bigg|\frac{p(\textbf{c}_{k+1,0}-\textbf{c}_{k,0})}{\sum_{z=\text{max}(1,k-p+1)}^k h_z}\bigg| 
    \leq \textbf{v}^{\text{max}}
    \, & k=1,..C \\
    |\textbf{c}_{k,2}|= & 
    \bigg|\frac{(p-1)(\textbf{c}_{k+1,1}-\textbf{c}_{k,1})}{\sum_{z=\text{max}(1,k-p+2)}^k h_z}\bigg| 
    \leq \textbf{a}^{\text{max}}
    \, & k=1,..C-1 \\ 
    |\textbf{c}_{k,3}|= & 
    \bigg|\frac{(p-2)(\textbf{c}_{k+1,2}-\textbf{c}_{k,2})}{\sum_{z=\text{max}(1,k-p+3)}^k h_z}\bigg| 
    \leq \textbf{j}^{\text{max}} \, & k=1,..C-2. \\
\end{aligned}
\end{equation}
Note that this condition is sufficient to ensure the validity of the constraints. Additionally, the optimisation variables have a lower bound: 
\vspace{-0.5cm}
\begin{equation}
    h_g^{\text{lb}}=\max_{\substack{i=\{x,y,z,\phi,\vartheta,\psi\}}} \bigg\{\frac{|w_{g+1}^i-w_{g}^i|}{v^{\text{max}}_i}\bigg\},
\end{equation}
since any interval between a pair of consecutive waypoints ($w_g^i, \, w_{g+1}^i$) cannot be run at infinite velocity.

\begin{figure}[!t]
    \centering
    \includegraphics[width=0.8\linewidth]{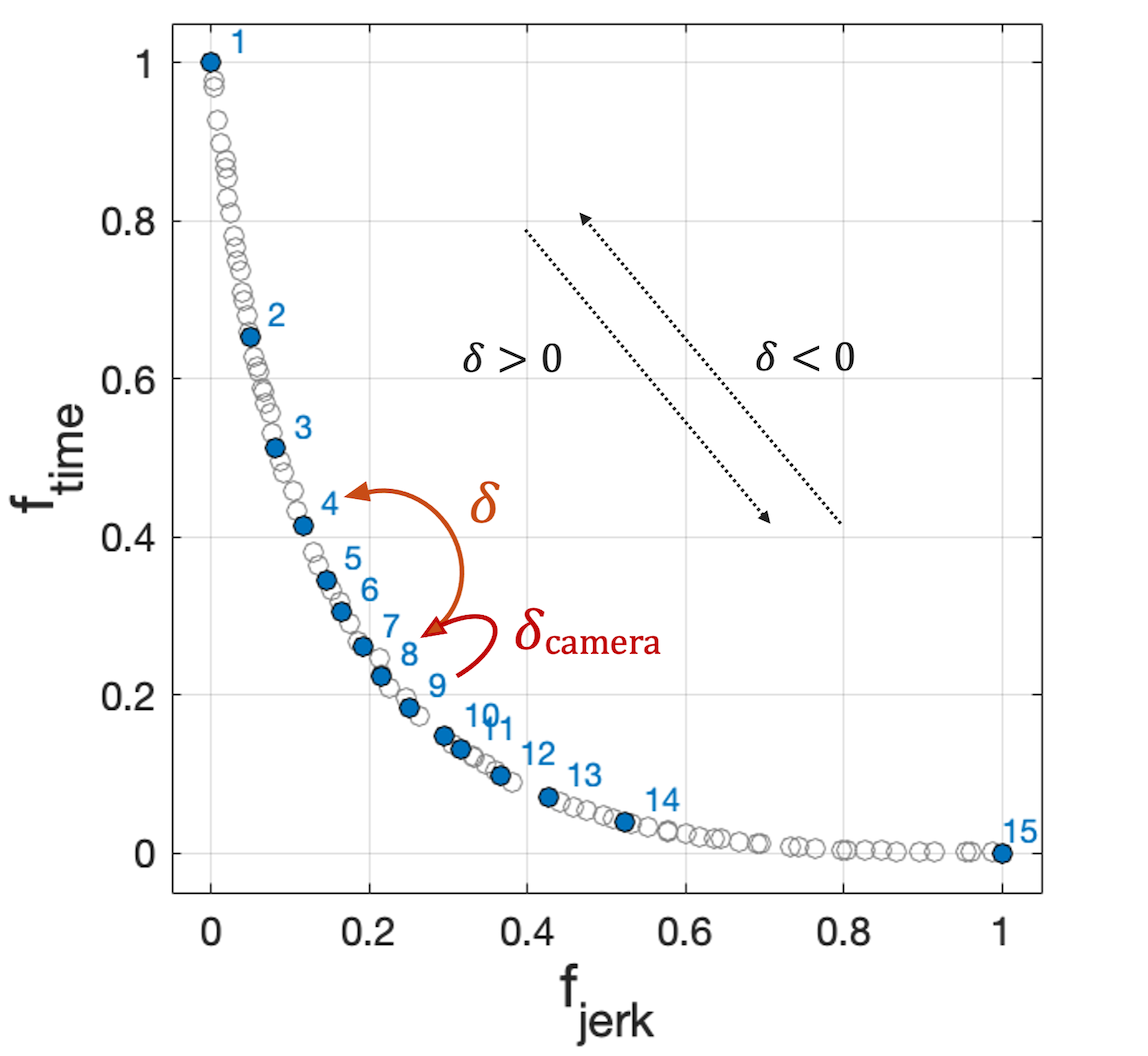}
    \vspace{-0.2cm}
    \caption{Pareto Optimal Front $\mathcal{F}$ (gray circles, normalised for sake of clarity) resulting by minimum time-jerk optimisation. Blue dots denote the front $\mathcal{F}^*$ after the downsampling procedure.}
    \label{fig:pareto}
    \vspace{-0.2cm}
\end{figure}

The constrained bi-objective problem results in a set of non-dominated solutions in the objective space, meaning no objective function can be improved without deteriorating the other objective. This set of solutions is known as the Pareto optimal front $\mathcal{F}$ and is represented by grey circles in \autoref{fig:pareto}. Each solution in $\mathcal{F}$ corresponds to a different interval vector \textbf{h} in the parameter space, resulting in CoBot trajectories with distinct characteristics. Throughout the paper, we refer to the resulting set of interval vectors as $\mathcal{X}$.

\subsection{Online HRV- and Camera-based Human Stress Monitoring}
\label{sec:stress}

An online decision-making algorithm is implemented to select the most appropriate interval vector $\bf{h}$ from the set $\mathcal{X}$ based on the human psycho-physical stress. The stress level is continuously monitored during the interaction by analysing the HRV and extracting behavioural information from camera images. 

HRV is a sensitive physiological condition marker reflecting the balance between the sympathetic and parasympathetic nervous systems.  
The analysis of HRV relies on the computation of the time intervals (also referred to as RR intervals) between two consecutive heartbeats (i.e. R peaks) and their variation over time. Among various metrics proposed in the literature, the mean RR interval was selected since it has been shown to be a reliable and sensitive measure of stress, even in ultra-short recordings ($10$-$30$\si{\second}) \cite{Nussinovitch2011}. 
In this study, we periodically compute the mean value $\overline{\text{RR}}_i$ within a window of fixed time duration (at iteration $i$) and then compare this value with the one at the previous iteration $\overline{\text{RR}}_{i-1}$. 

$\overline{\text{RR}}$ analysis provides a reliable estimate, but it may delay stress detection since it is updated within a specific time interval and relies on the body's increased physiological response.
To anticipate stress detection, we employ visual data analysis and calculate the camera-based \textit{stress level} score $\varrho \in [0,1]$ proposed in our previous paper \cite{Lagomarsino2022anonline}. Our approach examines body language online and, specifically, compares upper-body kinematics to baseline joint movements, as well as detects self-touching occurrences. If $\varrho$ surpasses a predefined threshold $\varrho_\text{th}$, we adjust the trajectory by moving within the solution space $\mathcal{F}$ with a step size of $\delta_\text{camera} = - 1$. $\delta_\text{camera}$ is zero otherwise. This adjustment aims to achieve a slightly smoother and slower trajectory, thereby mitigating the impact of stress on the human.

\begin{figure}[!b]
    \centering
    \vspace{-0.3cm}
    \includegraphics[width=\linewidth]{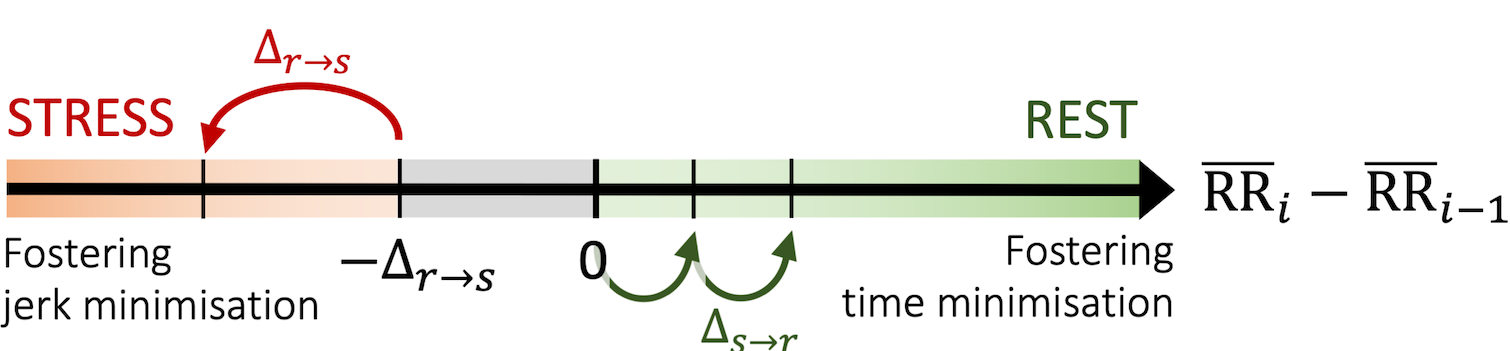} 
    \caption{Detection of changes in psycho-physical stress in relation to $\overline{\text{RR}}_i$ variations.}
    \label{fig:rr_variations}
\end{figure}

Intense psycho-physical stress is associated with a decrease of $\overline{\text{RR}}$. When the variation exceeds a minimum value $\Delta_{r\rightarrow s}$ and $\overline{\text{RR}}_i$ is lower than the baseline value at rest  $\overline{\text{RR}}_r$, i.e. 
\vspace{-0.1cm}
\begin{equation}
\label{eq:stress}
    (\,\overline{\text{RR}}_i -\overline{\text{RR}}_{i-1} <-\Delta_{r\rightarrow s}\,) \wedge  (\,\overline{\text{RR}}_i<\overline{\text{RR}}_r\,),
\vspace{-0.1cm}
\end{equation}
a change toward a higher stress level is detected.
In such a condition, we aim to select an interval vector from the set $\mathcal{X}$ that prioritises the smoothness of the desired CoBot trajectory over performance, giving more importance to minimising jerk than total execution time. Taking into account any previous adjustments made due to camera-detected stress, we move in the objective space solutions $\mathcal{F}$ of a number of steps: 
\vspace{-0.1cm}
\begin{equation}
\label{eq:step2stress}
    \delta = \Bigl\lfloor\dfrac{\Delta{\overline{\text{RR}}_r}+\Delta_{r\rightarrow s}}{\Delta_{r\rightarrow s}}\Bigr\rfloor + \delta_\text{camera} <0,
\vspace{-0.05cm}
\end{equation}
proportional to the variation of $\overline{\text{RR}}$ between consecutive time windows:
\vspace{-0.4cm}
\begin{equation}
    \Delta{\overline{\text{RR}}_r} = \overline{\text{RR}}_i -\min(\overline{\text{RR}}_{r}, \, \overline{\text{RR}}_{i-1}).
\vspace{-0.1cm}
\end{equation}
Note that if $\overline{\text{RR}}_{i-1}$ is above the value at rest $\overline{\text{RR}}_r$, the variation $\Delta{\overline{\text{RR}}_r}$ is performed with respect to resting conditions. Also, the floor function in \autoref{eq:step2stress} ensures that a slight variation in the stress range is mapped into a step in the solutions, while $N$ steps are computed with $\Delta{\overline{\text{RR}}_r}=-(N\!+\!1)\Delta_{r\rightarrow s}$. 

Since our final aim is to enhance productivity without perilously increasing workers' psycho-physical stress, positive $\overline{\text{RR}}$ variations out of the range of stress (see \autoref{fig:rr_variations}), i.e. 
\vspace{-0.1cm}
\begin{equation}
    (\, \overline{\text{RR}}_i -\overline{\text{RR}}_{i-1}>0) \wedge  (\,\overline{\text{RR}}_i\geq\overline{\text{RR}}_s\,),
\vspace{-0.1cm}
\end{equation}
are mapped into steps in the Pareto optimal front $\mathcal{F}$ toward faster but less smooth trajectories. In particular, $\delta$ is defined as: 
\vspace{-0.2cm}
\begin{equation}
    \label{eq:step2rest}
    \delta = \Bigl\lceil\dfrac{\Delta{\overline{\text{RR}}_s}}{\Delta_{s\rightarrow r}}\Bigr\rceil >0
\end{equation} 
where $\Delta{\overline{\text{RR}}_s}$ is here computed in relation to the stressful condition, i.e. 
\vspace{-0.1cm}
\begin{equation}
    \Delta{\overline{\text{RR}}_s} = \overline{\text{RR}}_i -\max(\overline{\text{RR}}_{s}, \, \overline{\text{RR}}_{i-1}),
\vspace{-0.1cm}
\end{equation}
to avoid pushing productivity when the subject is already stressed.
Note that $\Delta_{s\rightarrow r}>\Delta_{r\rightarrow s}$, reflecting the hysteresis of the human body in transitioning between different psycho-physical states \cite{Villani2018}. This means that the minimum variation required to detect a change in the psycho-physical state from rest to stress $\Delta_{r\rightarrow s}$ is higher than the transition between stress to rest $\Delta_{s\rightarrow r}$. 
Thus, we foster the minimisation of the execution time resulting in trajectories with higher values of kinematic quantities, while still respecting the upper bounds imposed by the optimisation inequality constraints. 
Furthermore, the ceil function in \autoref{eq:step2rest} ensures that even minor positive variations $\Delta{\overline{\text{RR}}_s}$ are mapped into a step towards faster solutions. Additional steps are performed for every multiple of $\Delta_{s\rightarrow r}$. 

Finally, no action is taken ($\delta\!=\!0$) for variations in $\overline{\text{RR}}$ within the range $[-\!\Delta_{r\rightarrow s}, 0]$. However, to prevent cumulative stress, we check whether $\overline{\text{RR}}_i$ has not gradually reached undesirable low values during the task. On that occasion, a step toward more relaxing trajectories is performed ($\delta\!=\!-1$).

\subsection{Trajectory Generation}
\label{sec:trajectory_generation}

\begin{figure}[!b]
    \vspace{-0.3cm}
    \centering
    \includegraphics[height=0.47\linewidth]{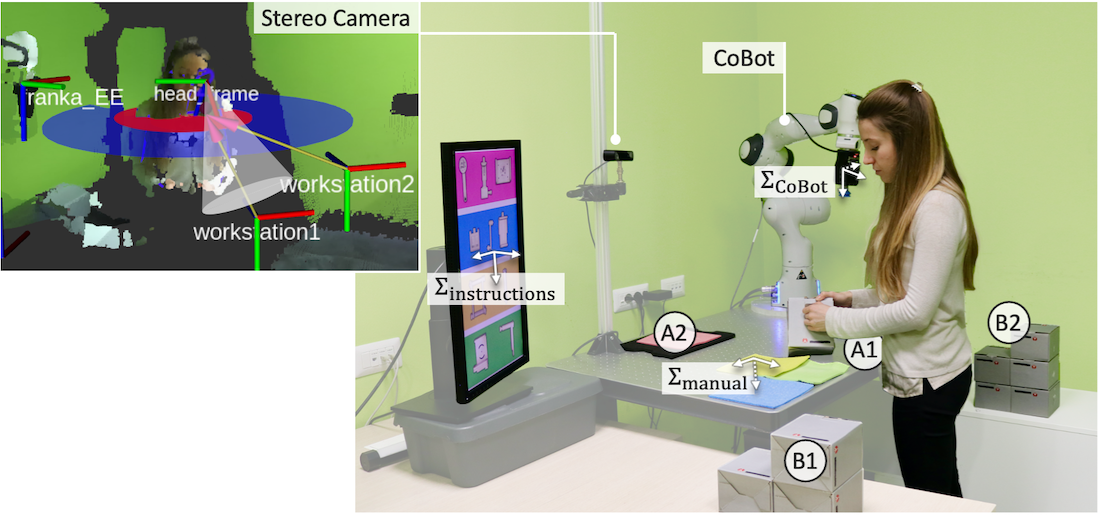}
    \caption{Setup of the \textit{Path Adaptation Experiment} including: human operator, CoBot, stereo camera, instructions monitor, and shared workplace.}
    \label{fig:exp1}
\end{figure}

Upon selecting the optimal \textbf{h}, we solve the interpolation problem for each component of the CoBot end-effector pose $\textbf{x}_\text{EE}$ by defining a system of $6+W$ equations. 
The six boundary conditions are enforced by equating the initial and final values assumed by $\dot{\textbf{x}}_\text{EE}$, $\ddot{\textbf{x}}_\text{EE}$, and $\dddot{\textbf{x}}_\text{EE}$ to the first and last control points of the corresponding B-spline defining the kinematic value along the trajectory (e.g. $\textbf{c}_{1,0}=\dot{\textbf{x}}_\text{EE}(0)=$ initial velocity). 
This is made possible by the fact that the basis function $N_{1,p-d}$ in the first knot $\tau_1$ has a value of $1$, while all the other basis functions $N_{k,p-d}(\tau_1)$ with $k=2,\dots,C+1$ are equal to $0$.
Similarly, in the last knot $\tau_{m+1}$, the basis function $N_{C+1,p-d}$ is the only non-zero function and has a value of $1$.
The remaining $W$ equations impose the passage through the waypoints (e.g. ${\textbf{x}_\text{EE}}(\tau_{(p+1)+3})=\textbf{w}_3$). 
All the equations can be expressed as a combination of the control points defining the end-effector trajectory: 
\vspace{-0.2cm}
\begin{equation}
    {\bf{\Theta}}_i=
    \begin{bmatrix}
        c_{1,0}^i\\
        \dots \\
        c_{C+1,0}^i
    \end{bmatrix}, \quad \text{with } \, i = \{x,y,z,\phi,\vartheta,\psi\}.
\vspace{-0.1cm}   
\end{equation}
Consequently, the system can be reformulated as the linear system:
\vspace{-0.1cm}
\begin{equation}
\label{eq:lin_system}
    {\bf{A}}\,{\bf{\Theta}}_i = {\bf{B}}_i, 
\end{equation}
where the matrix \textbf{A} is determined solely by the interval vector \textbf{h} and is identical for all elements of $\textbf{x}_\text{EE}$, while the unknowns correspond to the sets of control points ${\bf{\Theta}}_i$.

Utilising the optimal \textbf{h} obtained through the time-jerk optimisation and decision-making procedure (as discussed in previous sections), the matrix \textbf{A} is computed and employed in \autoref{eq:lin_system} to determine the control points for each pose component. Hence, the optimal Cartesian trajectory is completely defined. We evaluate the desired end-effector pose and velocity using \autoref{eq:bspline} and \ref{eq:velocity}, respectively and use them as references for a low-level Cartesian impedance controller.

\section{Experiments}
\label{sec:experiments}

\begin{figure}[!b]
    \vspace{-0.3cm}
    \centering
    \includegraphics[height=0.47\linewidth]{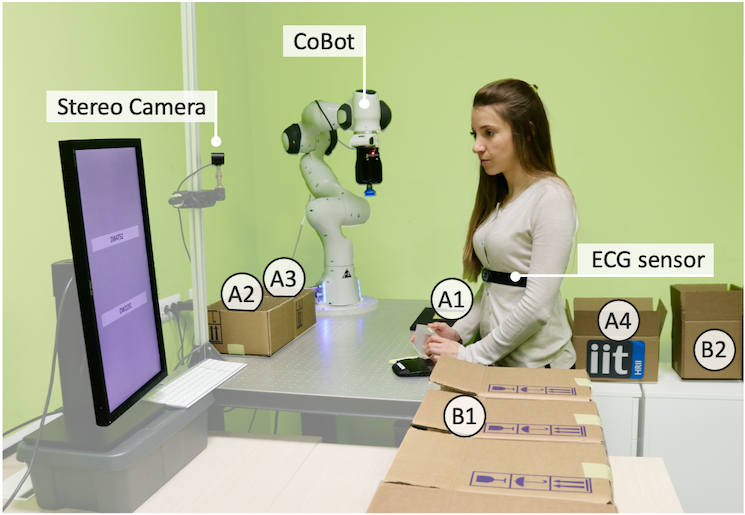}
    \caption{\textit{Overall Validation Experiment} enabling CoBot path and timing law adaptation and thus involving camera- and ECG-based human state monitoring.}
    \label{fig:exp2}
\end{figure}

The proposed architecture was tested with multi-subject experiments in two case studies, where the human and the CoBot worked side-by-side to achieve a shared goal. 
The tasks exemplified industrial logistics activities where the CoBot performed repetitive actions while human intervention allowed product customisation assuming to meet task requirements and customer demands. These cyclic activities were specifically designed to impose a mild mental demand on the operator and provide an environment conducive to developing cumulative work-related stress and time pressures. 
Before proposing the multi-subject experiments, we simulated the motion of both human right and left hands to showcase the robot's path adaptation scalability to multiple human body parts. To this aim, we focused on the adaptation based on purely physical factors considering the situation where the human is fully attentive to the robot's motion, with null mental effort  ($d_\text{physical}=d_\text{cognitive}=0.25\si{m}$).
In the first multi-subject experiment, referred to as \textit{Path Adaptation Experiment}, we validated the path adaptation strategy and compared its performance in terms of time and fluency with state-of-the-art approaches. 
The second experiment, denoted as \textit{Overall Validation Experiment}, was conducted to test the overall pipeline described in the previous sections. 
A video of the experiments can be found in the multimedia attachment or at \href{https://youtu.be/A3_0TwDlBLY}{\textbf{https://youtu.be/A3\_0TwDlBLY}}.

\subsection{Experimental Setup}

The setups of the \textit{Path Adaptation Experiment} and \textit{Overall Validation Experiment} are shown in \autoref{fig:exp1} and \ref{fig:exp2}, respectively. 
The participants stood in front of a table, where they were given a manual activity to perform (defining the region of interest $\Sigma_\text{task}$), and instructions for completing the collaborative tasks were displayed on a monitor (i.e. $\Sigma_\text{instructions}$). 
The tasks were carried out in collaboration with the Franka Emika Panda manipulator (controlled at $1$ kHz) equipped with the Robotiq’s vacuum gripper EPick. The end-effector frame defined the CoBot attention area $\Sigma_\text{CoBot}$. 

Throughout both experiments, a stereo camera (zed2, Stereolabs) monitored the subject from the front. The camera-based assessment of \textit{mental effort} and \textit{stress level} ran at around $20$ Hz, and the adaptive safety zones were consequently adjusted based on the current distance between the centre of the $3$D bounding box surrounding the human and the CoBot end-effector. 
Utilising the bounding box allows considering all skeleton keypoints together while accounting for user arms's motion away from the torso.
The distance thresholds $d_\text{collision-free}$, $d_\text{conservative}$, $d_\text{social}$ were set to $0.25$, $0.40$, $1.00$\si{\meter}, as suggested in \cite{kirschner2022expectable}. Moreover, we experimentally define the span parameter $\gamma = 0.4$, motion divergence $\beta_\text{th} = \pi/3$, and \textit{stress level} threshold $\varrho_\text{th} = 0.5$.

During the \textit{Overall Validation Experiment}, the ECG signal of the participant was also monitored by Polar H10 chest strap ($130$ Hz). Cardiovascular data were streamed via Bluetooth to the CPU running the proposed framework and processed to extract RR intervals. Every $30$ seconds, the mean value of inter-beats intervals $\overline{\text{RR}}_i$ in the past window was computed and used to enhance the collaboration. The architecture and communications were implemented using Robot Operating System (ROS) platform.

The CoBot followed trajectories generated by interpolating a sequence of waypoints with the proposed quintic B-spline trajectory planner, in charge of sending suitable references to a low-level Cartesian impedance controller. The kinematic quantities at the boundaries were set to zero to facilitate the pick-and-place operations involved in the task. The upper bounds were determined based on the specifications provided in the CoBot datasheet\footnotemark.
The constrained bi-objective optimisation problem was solved using pymoo \cite{pymoo} and the NSGA-II algorithm with a population size of $90$. The resulting Pareto front $\mathcal{F}$ and its corresponding solution set $\mathcal{X}$ were downsampled, as shown in \autoref{fig:pareto}. 
To determine the final samples and parameters of the stress-based decision-making algorithm, we relied on \cite{Villani2018}. We set $\Delta_{r\rightarrow s}\!=\!0.02$\si{\second}, $\Delta_{s\rightarrow r}\!=\!0.01$\si{\second} and assumed that $\overline{\text{RR}}$ could vary within the range $[\overline{\text{RR}}_s\!-\!\sigma_s, \overline{\text{RR}}_r\!+\!\sigma_r]$, where $\sigma_s$, $\sigma_r$ represent the standard deviations observed in \cite{Villani2018} in stress and rest conditions.
Thus, $\overline{\text{RR}}_s^\text{max}\!\!\!=\!\overline{\text{RR}}_s\!-\!\sigma_s$ and we determined the number of solutions in the downsampled set $\mathcal{X}^*$ as $\frac{(\overline{\text{RR}}_r+\sigma_r)-(\overline{\text{RR}}_s-\sigma_s)}{\Delta_{r\rightarrow s}}\!\approx\!15$. Augmented Scalarization Function decomposition method was exploited to seek fifteen solutions out of $\mathcal{X}$, recursively assigning slightly more importance to the objective function $f_\text{time}$ than to $f_\text{jerk}$. As a result, the first element of $\mathcal{X}^*$ was the interval vector $\textbf{h}^1$ resulting in the minimum-jerk trajectory, while the last, i.e. $\textbf{h}^{15}$, defined a minimum-time trajectory. Solutions in between represented trade-offs between these two objectives. Selecting a higher solution number resulted in a shorter cycle execution time. 

\footnotetext{Cartesian trajectory requirements: 
\href{https://rb.gy/cctqy}{https://rb.gy/cctqy}}

\subsection{Experimental Protocol}

Twenty-four healthy participants, thirteen males and eleven females ($28.2 \pm 3.0$ years old), were recruited from the students and research personnel of Istituto Italiano di Tecnologia\footnote{Experiments were carried out at Human-Robot Interfaces and Interaction (HRII) Lab, Istituto Italiano di Tecnologia (IIT), in accordance with the Helsinki Declaration, and the protocol was approved by the ethics committee Azienda Sanitaria Locale Genovese N.3 (Protocol IIT\_HRII\_ERGOLEAN 156/2020).}.
\footnotetext[5]{\url{https://www.kubios.com}}

The \textit{Path Adaptation Experiment} (\autoref{fig:exp1}) asked the participant to pick a box from location $B1$. The box contained an assembly and the subject had to identify it from a list of products displayed on the monitor, select the associated inkwell, and place it in $A2$. The CoBot took the ink from $A2$ and stamped it onto the box placed by the user in $A1$. Finally, the participant retrieved the stamped box and put it in $B2$ for dispatch. This activity was repeated for ten boxes. The participant was free to move around and intrude into the CoBot workspace, while the proposed architecture continuously adapted the path to ensure physical and cognitive-grounded safety zones around the human. Note that the CoBot held a fixed timing law for the whole duration of this experiment.

The \textit{Overall Validation Experiment} (\autoref{fig:exp2}) consisted in sorting the items for recycling. The subject received a box (in $B1$) containing objects of different materials. Each object was labelled with a code, and on the monitor, the items that had to be recycled were specified. The operator had to provide the CoBot with them in $A1$. The robotic assistant was responsible for sorting them according to the material and placing them in $A2$ or $A3$. Spare items had to be thrown away in $A4$ and the empty box placed in $B2$.
Meanwhile, a new box arrived, and a new cycle could begin.

\begin{figure}[!t]
    \centering
    \includegraphics[width=\linewidth]{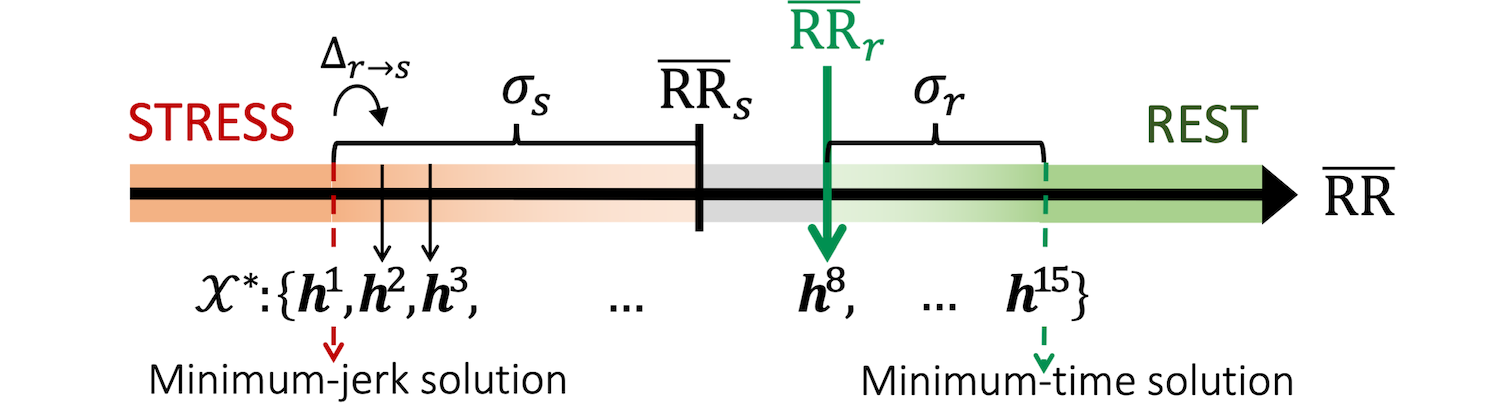}
    \vspace{-0.4cm}
    \caption{Initial mapping between $\overline{\text{RR}}$ intervals extracted by ECG signal and downsampled optimal solution set $\mathcal{X}^*$.
    \vspace{-0.3cm}
    }
    \label{fig:initial_mapping}
\end{figure}

Before the start, we recorded the person-specific baseline values of ECG and measured the mean inter-beat interval during resting conditions, referred to as $\overline{\text{RR}}_r$. 
Additionally, the participants performed a training task to familiarise themselves with the setup and capture the upper-body joint kinematics at rest, which was used later as a reference for calculating the camera-based \textit{stress level} score.
The experiment employed a within-subjects testing design in which each subject underwent three distinct robot control strategies, each involving the completion of ten boxes. 
The order was randomised, and breaks were provided between sessions to mitigate potential learning effects and cumulative workload. In condition \textit{a} and \textit{b}, the CoBot adhered to fixed-path minimum-time and minimum-jerk trajectories, respectively, throughout the entire session duration. Conversely, condition \textit{c}, which constituted the main focus of this experimental analysis, involved human-aware adjustment of CoBot trajectory both in terms of path and associated timing law. 
\autoref{fig:initial_mapping} illustrates the initial mapping between the HRV feature and $\mathcal{X}^*$. 
The timing law adaptation algorithm started from solution $\textbf{h}^8$ (i.e.  $15-\frac{\sigma_r}{\Delta_{r\rightarrow s}}\!\!\approx\!8$) assuming that the participant began the experiment in a resting state. 
It is worth noting that the mapping is personalised to each individual, considering their specific baseline measured during the calibration phase. 

\subsection{Comparison with State-of-the-Art Approaches} 
We compared our method for ensuring safety with two State-of-the-Art (SoA) methods resulting in static safety zones: a straightforward Velocity Scaling (VS) approach as in \cite{Palleschi2021} and the Expectable Motion Unit (EMU) proposed in \cite{kirschner2022expectable}. 
To this aim, we conducted simulations using the human motion data collected during multi-subject experiments. By running the SoA method on this data, we were able to assess different strategies under identical conditions. 

\subsubsection{Velocity Scaling}
In VS, the attention-based procedure outlined in \autoref{sec:path_adaptation} is replaced by a module that, given the planned velocity $\Dot{\textbf{x}}_\text{EE}$, generates a scaled velocity $\Dot{\textbf{x}}_\text{EE}^\text{scaled} = \delta \, \Dot{\textbf{x}}_\text{EE}$ at each iteration. 
When the measured human-robot separation distance $d(t)$ falls within the range $[d_\text{collision-free}, d_\text{social}]$, an upper bound for the end-effector speed, denoted as $v_\text{VS}(t)$, is computed using the increasing bell-shaped function:
$$\quad \quad v_\text{VS}(t) = \dfrac{0.250}{2}\cos{\Big[\dfrac{(d{\scriptstyle (t)}-d_\text{collision-free})\pi}{d_\text{social} - d_\text{collision-free}} +\pi\Big]},$$ 
where $0.250$\si{\meter/\second} is the limit for safe speed control specified in the ISO-10218-1.
Consequently, the scaling factor is calculated as $\delta(t) = v_\text{VS}/|\Dot{\textbf{x}}_\text{EE}|$. Note that $\delta$ is set to $0$ if $d < d_\text{collision-free}$, and $\delta = 1$ if $d > d_\text{social}$.

\subsubsection{Expectable Motion Unit}
We exploit the model of rapid Involuntary Motion (IM) obtained in \cite{kirschner2022expectable} and the EMU linear mapping between separation distance and robot velocity $v_\text{EMU}(t)$ to restrict the probability of IM occurrence to $15\%$. Similarly to VS, the planned velocity $\Dot{\textbf{x}}_\text{EE}$ is limited when needed so that the end-effector speed does not exceed $v_\text{EMU}(t)$. Note that an offset equal to $0.25$\si{\meter} was added since we considered the distance from the centre of the bounding box surrounding the human instead of its border.

\subsection{Measurements and derived metrics}
This section presents the adopted measurements and derived metrics to assess the framework. Statistical analysis was conducted to compare our findings with those obtained from SoA methods and across testing conditions. The gathered data were first tested for normality using the Anderson-Darling test. Upon normality confirmation, paired t-tests were run. Otherwise, we employed the non-parametric Wilcoxon signed-rank test to assess any significant differences. 

\subsubsection{Collaboration Fluency} 
We adopted fluency metrics from \cite{scalera2022enhancing}. Specifically, we measured the task time $T$, which indicates the duration the robot took to complete the task, directly reflecting the system productivity. Besides. the robot idle time $T_\text{r-idle}$ represents the percentage of total task time during which the robot is not visibly active. 

\subsubsection{Human Mental Effort and Psycho-physical Stress} 

After the experiments, participants assessed the perceived workload in various testing conditions by completing the NASA Task Load Index questionnaire \cite{Hart1988}. Additionally, we asked whether the system met their needs concerning perceived safety and comfort and their performance demands. The subjects provided their feedback using a five-point Likert scale. In cases where they felt the need to specify particular concerns, they were encouraged to tick if the robot approached too closely or remained too distant, if its motion was jerky, excessively slow or fast, or if its path was perceived as too long or short. 

A detailed cardiovascular data analysis was conducted using Kubios software\footnotemark[5]. This tool calculates several standard metrics in the time, frequency, and non-linear domains. We examined the mean RR intervals to assess the psycho-physical stress induced by different robot control strategies during the collaborative task. Additionally, we explored the LF/HF ratio derived from the frequency domain HRV data, as it is considered an indicator of mental effort according to existing literature \cite{Durantin2014}. 
For inter-subjects comparisons, the HRV features were normalised based on each individual's baseline. 

\subsubsection{Key Performance Indicators}
The production rate of each operator $o$ under diverse conditions was evaluated in terms of completed boxes per minute:
\begin{equation}
    \hspace{0.3cm} \rho^{s}_o = \frac{60}{T^{s}/b^{s}_o},
\end{equation}
where $T^{s}$ denoted the total time required to complete  $b^{s}_o$ cycles of condition $s$ (\textit{a}, \textit{b} or \textit{c}).
To assess user performance, 
an error rate metric was defined as: 
\begin{equation}
    \epsilon^{s}_o = \frac{e^{s}_o}{b^{s}_o},
\end{equation}
indicating the errors committed per cycle.
An error $e^{s}_o$ was recorded whenever the participant failed to select objects with the correct code or did not place objects in time for the robot to pick them.
The average production and error rates across all participants were used to compare the testing conditions.

\section{Experimental Results}
\label{sec:result}

\subsection{Path Adaptation Experiment}

\subsubsection{Simulation}

\begin{figure}[!b]
    \centering
    \vspace{-0.3cm}
    \includegraphics[width=\linewidth]{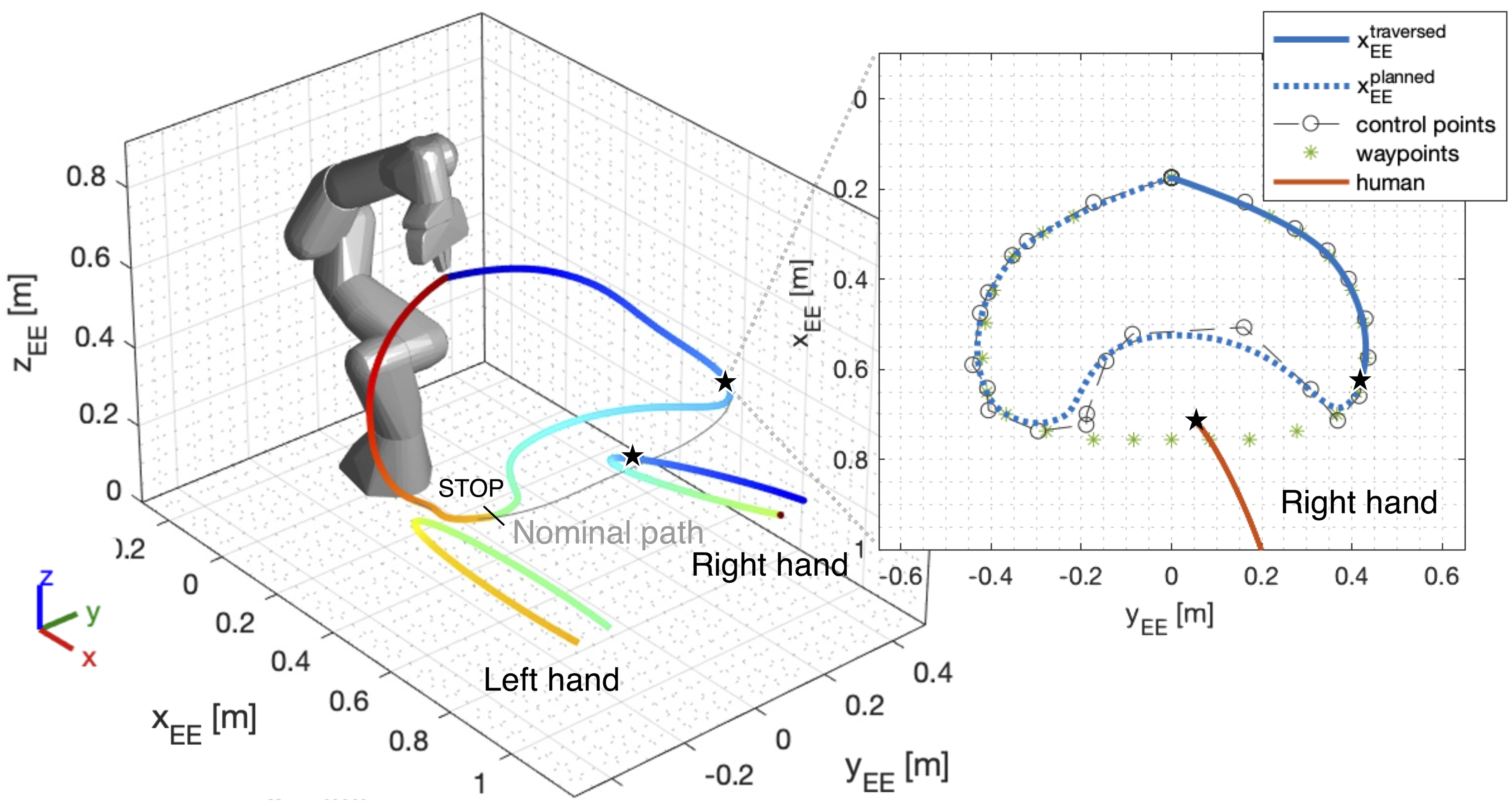}
    \vspace{-0.5cm}
    \caption{Visualisation of the robot trajectory execution to adapt to human hands' movements. Close-up of traversed robot path up to the marked time instant (black star) and planned adaptations of control points, polyline, and path based on the current motion of the right hand.}
    \label{fig:path_adapt_simulation}
\end{figure}

To illustrate the robot's path adaptation capability to multiple body parts, \autoref{fig:path_adapt_simulation} presents the outcome of a simulated scenario. As the human right hand entered the robot's workspace, the trajectory gradually adapted on the fly. In the close-up view, we provide details of this trajectory adaptation at the time marked by a black star. We display the traversed robot path (blue solid line) and human right-hand movement (orange solid line) up to this specific instant and include the planned adaptation of the control points (black spheres), the determined polyline (black dashed line), and the path (blue dotted line) to demonstrate the dynamic adjustments performed to complete the trajectory. Subsequently, as can be seen in the left plot, a simulated sudden and unexpected movement of the left hand towards the robot resulted in a temporary halt, but the robot promptly resumed movement once the hand moved away sufficiently. It should be noted that the stop condition was not triggered if the robot path diverged from the motion of the human body parts by an angle greater than $\beta_\text{th}$.

\begin{figure}[!t]
    \centering
    \includegraphics[width=0.9\linewidth]{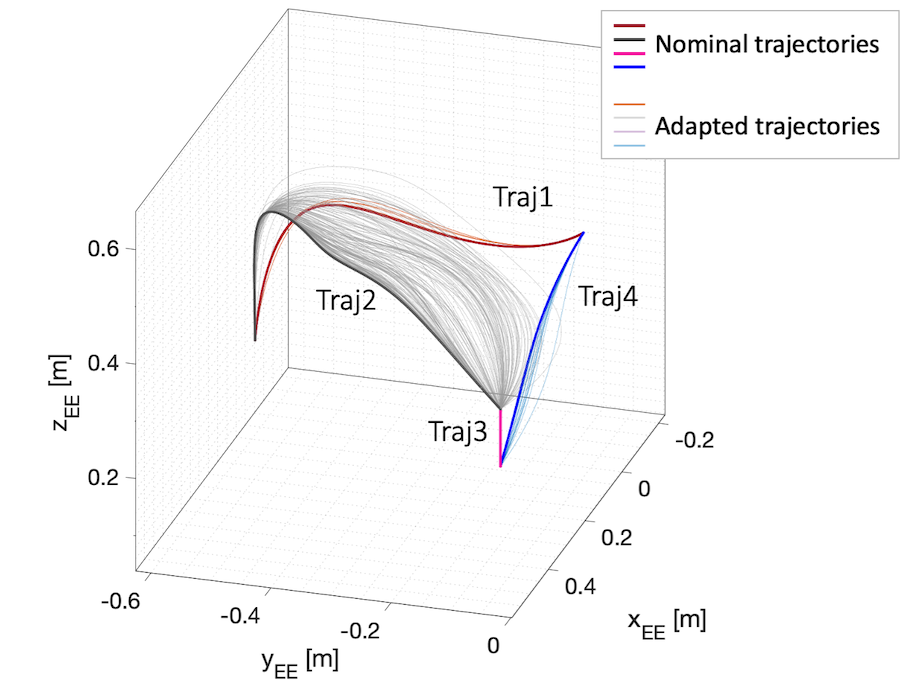}
    \vspace{-0.2cm}
    \caption{Results of path adaptation strategy over all participants.}
    \label{fig:variability}
    \vspace{-0.3cm}
\end{figure}

\begin{figure}[!b]
    \vspace{-0.3cm}
    \centering
    \begin{adjustwidth}{-0.1cm}{-0.3cm}
    \includegraphics[width=\linewidth]{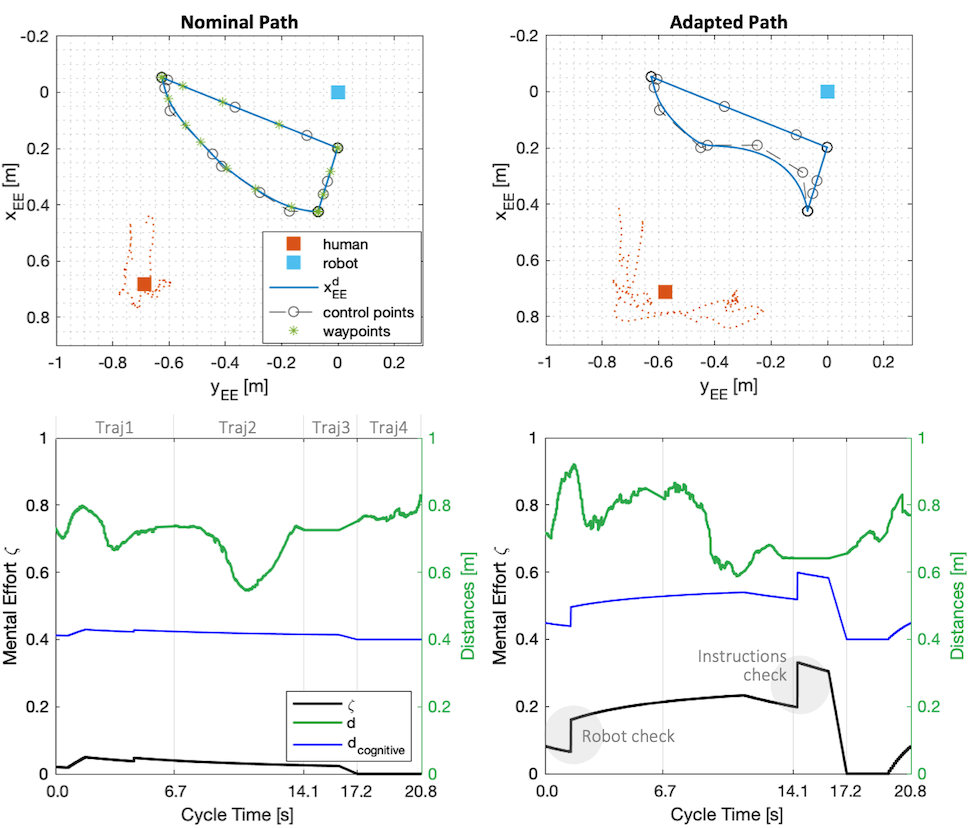}
    \end{adjustwidth}
    \vspace{-0.2cm}
    \caption{Online path adaptation to ensure cognitive-grounded safety based on estimated \textit{mental effort} experienced by a subject.
    }
    \label{fig:results_cognitive}
\end{figure}

\subsubsection{Method Validation}

\autoref{fig:variability} presents the results of the path adaptation strategy for all participants of the first multi-subject experiment. 
The thicker lines represent the nominal path of the CoBot end-effector, while the thinner lines depict the adaptations made. 

To provide further insights into the algorithm's functioning, we provide the results of a specific participant in \autoref{fig:results_cognitive}. In the top half of the figure, the blue line shows the projection of the end-effector path on the horizontal plane during a cycle of the activity. A blue square denotes the position of the robotic base. 
The movements of the human partner are depicted by a red dotted line, with the red square indicating the centroid over the entire cycle duration. On the left side of the figure, we present a task cycle where the estimated \textit{mental effort} (black line, graphs on the bottom) of the participant was low, and the separation distance between the human and the CoBot (green line) was above $d_\text{cognitive}(t)$ (blue line). As a result, the end-effector path remained unaltered. Conversely, in the cycle shown on the right, the participant experienced higher mental demand, leading to an increased radius of the cognitive-grounded safety zone $d_\text{cognitive}(t)$. This resulted in a deviation of the end-effector path to ensure a greater separation distance between the human and the CoBot.
What is striking about the approach is that the path alteration does not impact the continuity of the end-effector path, as can be seen in the figure, or cause any delay in the total execution time of the cycle.

\begin{figure}[!t]
    \centering
    \begin{adjustwidth}{-0.3cm}{-0.1cm}
    \includegraphics[width=\linewidth]{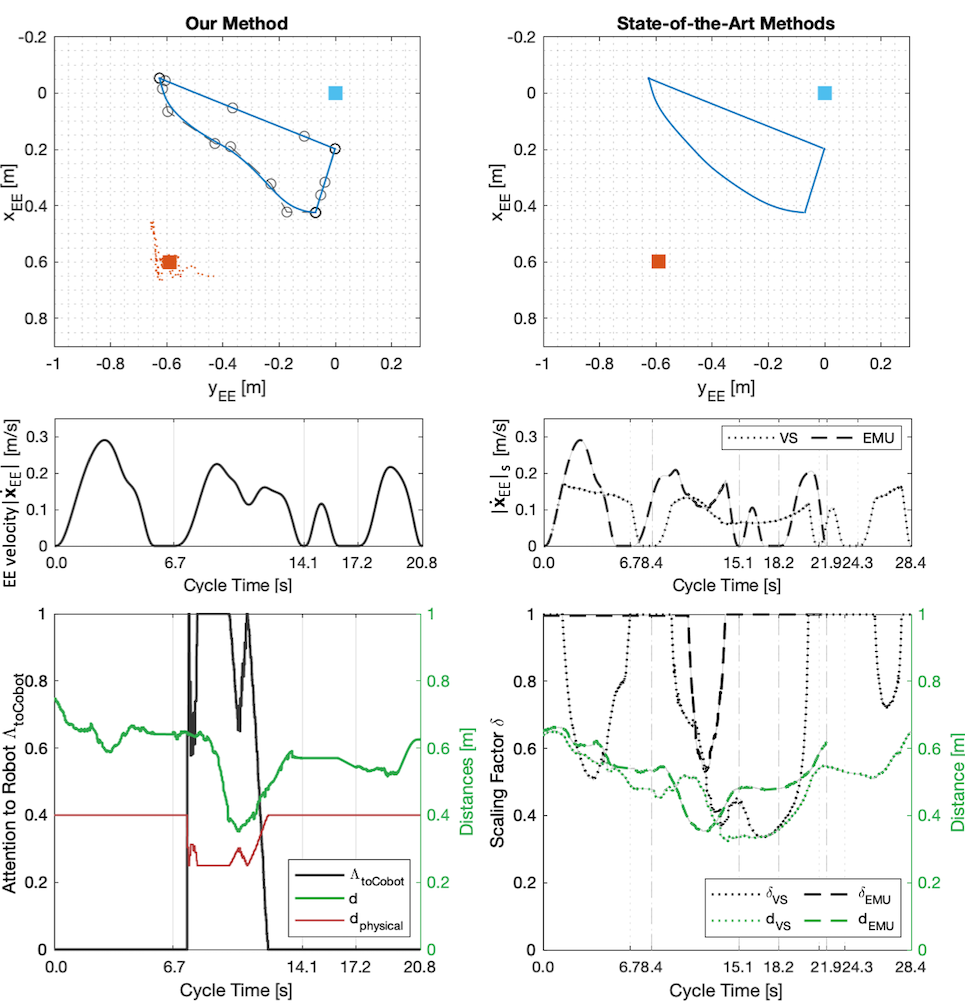}
    \end{adjustwidth}
    \vspace{-0.2cm}
    \caption{Comparison between our attention-based approach to guarantee physical safety and state-of-the-art methods.
    \vspace{-0.3cm}
    }
    \label{fig:results_physical}
\end{figure}

The proposed strategy for physical safety management is illustrated on the left side of 
\autoref{fig:results_physical}. 
When the human was aware of its motion, our method allowed the CoBot to approach closer without stopping. This was achieved by online adjusting the radius $d_\text{physical}$ of the physical safety zone. 
The adjustment of $d_\text{physical}$ (red line, graphs on the bottom) was based on the percentage of attention $\Lambda_\text{toCoBot}$ (black line) that the individual directed towards the CoBot. 

\subsubsection{Comparison with State-of-the-Art Approaches} 

\autoref{fig:results_physical} compares the outcomes of the proposed method with the existing literature for a participant. The figure reveals how the SoA approaches significantly reduced the CoBot speed (graphs in the middle where the dashed line refers to VS while the dotted line to EMU), affecting the total cycle time. 
\autoref{tab:results_fluency} reports the average fluency metrics obtained across all participants. Using VS and EMU methods, the presence of the human operator triggered the safety module, resulting in reduced speeds during the approach phases and periods of inactivity. The statistical analysis revealed a significant reduction in task time $T$ (the p-value obtained comparing with VS and EMU methods was $\text{p} < 0.001$), making it almost comparable to the scenario without a human ($\text{p} = 0.176$). The path adaptation and monitoring of human awareness made CoBot idle time almost null ($\text{p} = 0.002$ from the comparison with both VS and EMU methods, respectively). 
Furthermore, our approach successfully maintained a separation distance of at least $d_\text{collision-free}$ throughout all experiments. In contrast, the VS and EMU methods failed to achieve this for five and three participants, respectively. 

\begin{table}[!t]
    \centering
    \caption{Comparison of Fluency Metrics between our approach and state-of-the-art methods.}
    \label{tab:results_fluency}
    \resizebox{\linewidth}{!}{%
    \begin{tabular}{@{}lllll@{}}
    \toprule
     & $T$ {[\si{\second}]} & $T_\text{r-idle}$ {[\si{\second}]} & $d^\text{min}$ {[\si{\meter}]} & ${\bf{v}}^\text{mean}$ {[\si{\meter/\second}]} \\
     &
      {\color[HTML]{676767} mean(std)} &
      {\color[HTML]{676767} mean(std)} &
      {\color[HTML]{676767} mean(std)} &
      {\color[HTML]{676767} mean(std)} \\ \cmidrule(r){1-1} \cmidrule(r){2-3} \cmidrule(l){4-5} 
    Our Approach
        & 221.5 (11.9) {\color[HTML]{676767}\rdelim]{1.5}{1.5mm} \color[HTML]{676767}\rdelim]{3}{1.5mm} \multirow{2}{1mm}{**}} & 0.14 (0.79) \,{\color[HTML]{676767}\rdelim]{1.5}{1mm} \multirow{2}{0.5mm}{*}  \color[HTML]{676767}\rdelim]{3}{1.5mm} \multirow{3}{1mm}{*}} & 0.36 (0.04) & 0.133 (0.001)  \\
    VS
        & 283.5 (55.5) & 1.99 (0.23)  & 0.30 (0.06) & 0.106 (0.010)  \\
    EMU
        & 236.7 (17.2) & 6.60 (16.40) & 0.29 (0.06) & 0.129 (0.004)  \\
    Without Human
        & 220.8 & -  & - & 0.132                     \\ \bottomrule
    \end{tabular}
    }
    
    \smallskip
    Significance levels are indicated at the *$\text{p}<0.05$, **$\text{p}<0.001$. 
    \vspace{-0.3cm}
\end{table}

\begin{figure*}[!b]
    \includegraphics[width=\linewidth]{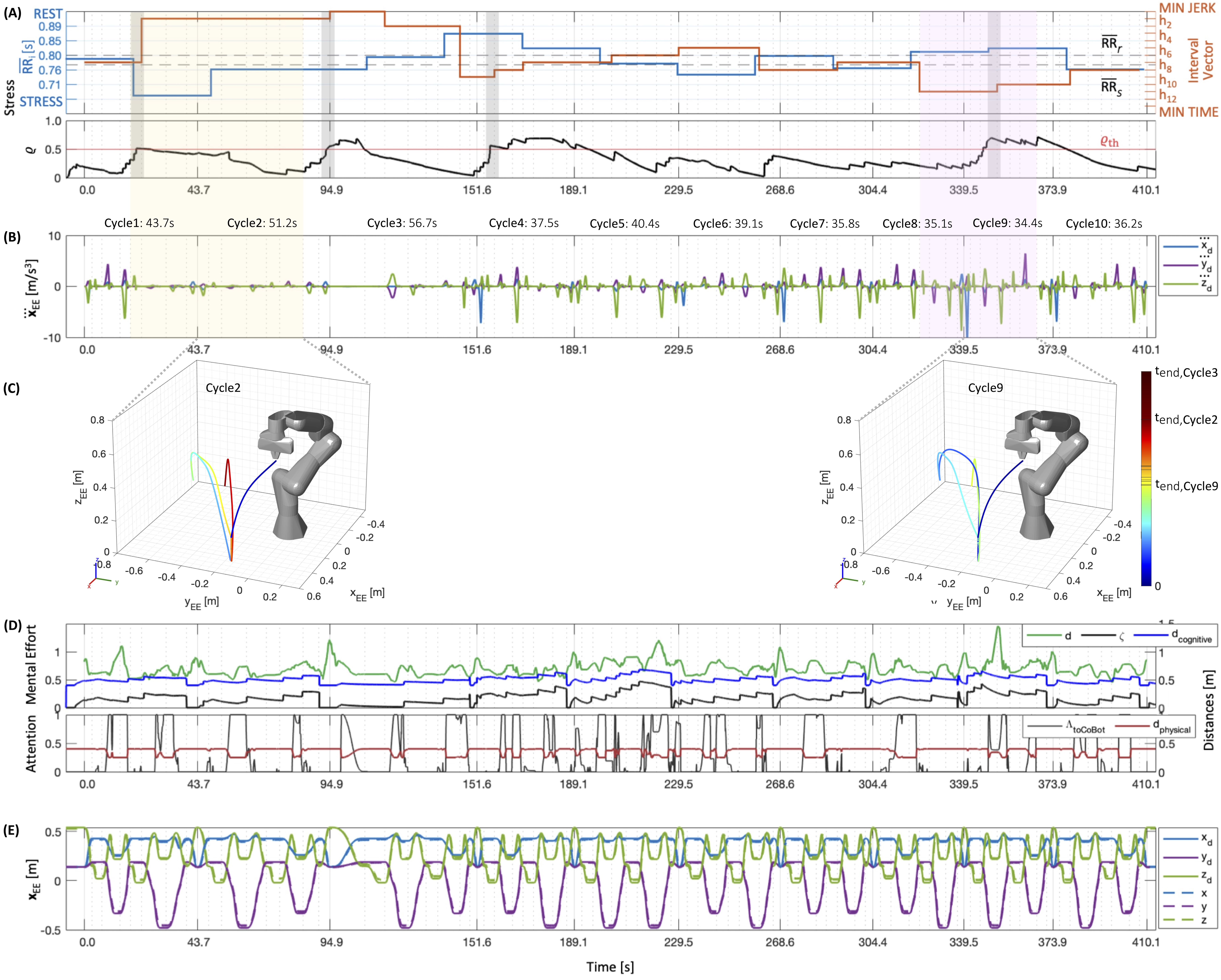}
    \vspace{-0.5cm}
    \caption{Excerpt of data from one subject collected during the \textit{Overall Validation Experiment}. 
    According to human stress monitored through $\overline{\text{RR}}$ and camera $\varrho$ \textbf{\textit{(A)}}, 
    the CoBot optimises its end-effector timing law resulting in varying cycle times and jerk values \textbf{\textit{(B)}}. When stress levels are high, the CoBot reduces jerk values and extends cycle times (shaded yellow area), while during relaxation periods, it exhibits more jerky movements and reduces cycle times (shaded pink area). Two plots in \textbf{\textit{(C)}} illustrate the highlighted cycles to better visualise the timings within the adapted paths.
    Indeed, based on human-robot distance $d$, attention to the CoBot $\Lambda_\text{toCoBot}$ and mental effort $\zeta$ \textbf{\textit{(D)}}, the CoBot simultaneously adapts its end-effector path, while maintaining continuity \textbf{\textit{(E)}}.}
    \label{fig:continuity}
\end{figure*}

\subsection{Overall Validation Experiment}

\subsubsection{Method Validation}

\autoref{fig:continuity} presents the multi-modal adaptation proposed by the PRO-MIND framework, reporting the data of an entire experiment (i.e. condition \textit{c}) for one of the twenty-four participants. In all the graphs (with time on the x-axis), the grey vertical solid lines delimit ten cycles (whose duration is also reported) of the performed collaborative object sorting, while dotted vertical lines mark the trajectories accomplished by the CoBot within a cycle. 

The online decision-making procedure to determine the optimal interval vector between consecutive waypoints based on HRV and camera-detected stress is shown in \autoref{fig:continuity}\textbf{\textit{A}}. 
The blue profile depicts the average value of RR intervals in $30$-second time windows, while the red line illustrates the selected solution. All participants started from a resting state, so the initial CoBot trajectory employed the eighth solution from the optimal set of interval vectors $\mathcal{X}^*$. 
The adaptation of the timing law associated with the executed CoBot trajectories resulted in varying cycle times and motion smoothness (see \autoref{fig:continuity}\textbf{\textit{B}}).
When an increase in psycho-physical stress was detected (i.e. a decrease in the blue profile, as indicated by the shaded yellow area), the system transitioned towards solutions that reduced the kinematic quantities (end-effector velocity, acceleration, and jerk), endorsing smoothness and protracted CoBot motion. Conversely, solutions resulting in fast but less smooth motions were selected when the user was relaxed (higher jerk values in the shaded pink area). 
The shaded grey areas indicate instances when stress was detected through the camera-based assessment method, i.e. moments when the score $\varrho$, represented by the black line, exceeded $\varrho_\text{th} = 0.5$. This detection promptly triggered a step adjustment in the optimal solutions toward smoother and slower trajectories. 
 
The two plots presented in Figure 13\textbf{\textit{C}} provide a clearer visualisation of the disparity in cycle execution time between the depicted cycles (also highlighted by differences in $t_\text{end}$ values in the colorbar). It is worth noting that the trajectories followed by the CoBot in these cycles exhibited variations not only in timing but also in the paths traversed. 
Indeed, the monitoring of human-robot separation distance $d$, mental effort $\zeta$, and attention to the CoBot $\Lambda_{\text{toCoBot}}$ enabled scaling safety zones around the user (see \autoref{fig:continuity}\textbf{\textit{D}}) and thus adapting the end-effector path. 
Finally, \autoref{fig:continuity}\textbf{\textit{E}} depicts the CoBot end-effector positions throughout the session (dotted and solid lines refer to desired and actual positions, respectively), demonstrating that trajectory continuity and smoothness remained unaffected despite trajectory adaptation. This is also evident from the outline of the end-effector jerk in \autoref{fig:continuity}\textbf{\textit{B}}.

\subsubsection{Human Mental Effort and Psycho-physical Stress}

\begin{figure*}[!b]
    \vspace{-0.1cm}
    \centering
    \includegraphics[width=\linewidth]{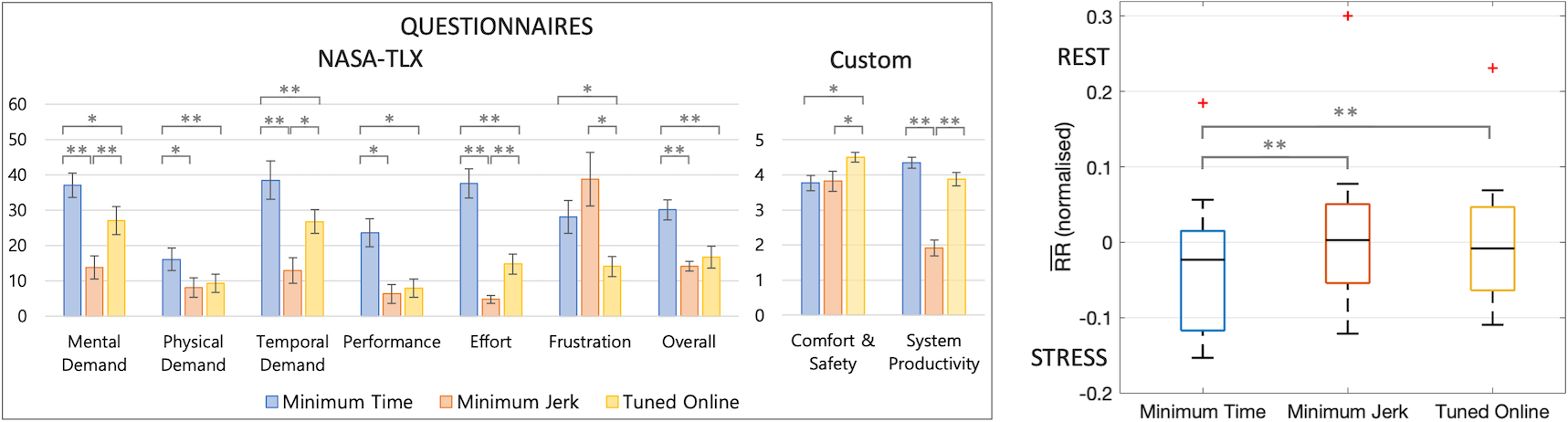}
    \vspace{-0.4cm}
    \caption{Left: Results of NASA TLX and custom questionnaire: bars represent mean and standard error between subjective ratings assigned by participants to different experimental conditions. Right: Normalised $\overline{\text{RR}}$ intervals during testing conditions. Significance levels are indicated at the *$\text{p}<0.05$, **$\text{p}<0.001$.}
    \label{fig:questionnaires}
\end{figure*}

\autoref{fig:questionnaires}(left) shows the results of subjective rating scales in the three experimental conditions. The bars denote the mean values assigned by participants for the NASA TLX and custom questionnaire items, while the error bars indicate the 95\% confidence interval of the scale means. 
The statistical analysis revealed a significant reduction in the mental demand (p-value between condition \textit{a} and \textit{b} was $\text{p}_{a,c} = 0.013$), perceived effort ($\text{p}_{a,c} < 0.001$), and temporal demand ($\text{p}_{a,c} < 0.001$) when utilising the trajectory adaptation strategy (condition \textit{c}) compared to fixed-path minimum-time trajectories (condition \textit{a}). At the same time, the proposed strategy maintained a higher level of human performance ($\text{p}_{a,c} = 0.010$). Interestingly, participants rated their own performance as comparable to the cosy condition where the CoBot follows minimum-jerk trajectories (condition \textit{b}). Even if a decrease in mental and temporal demand and effort was registered, the substantial limitation on CoBot Cartesian velocity of minimum-jerk trajectories resulted in excessive caution, leading to a noticeable level of human frustration ($\text{p}_{b,c} = 0.009$).
Overall, prioritising the time minimisation over the jerk in fixed-path CoBot trajectories significantly impacted the NASA-TLX workload score ($\text{p}_{a,b} < 0.001$). Nevertheless, by online adjusting the effects of the two optimisation objectives based on the current human psycho-physical stress and modifying the CoBot path on the fly, we managed to ensure an acceptable workload for the users, similar to the load experienced in condition \textit{b} ($\text{p}_{b,c} = 0.226$).

Notably, our human-in-the-loop path and timing low adaptation framework enhanced comfort and perception of safety ($\text{p}_{a,c} = 0.009$, $\text{p}_{b,c} = 0.028$). 
Among the participants, eight individuals specifically mentioned that the CoBot approached too closely in conditions \textit{a} and \textit{b}, where the path was not adapted. Only one participant perceived the path adaptation as resulting in an excessively long CoBot route. 
Furthermore, over half of the subjects described minimum-jerk CoBot trajectories as unnecessarily slow, while a similar proportion found minimum-time motions excessively fast. In contrast, only three participants perceived the time-jerk optimised motion as too slow, and one considered it too fast. 

Regarding the HRV analysis, the comparison of normalised $\overline{\text{RR}}$ values extracted from the human ECG signal during CoBot trajectories utilising different optimisation strategies is depicted in \autoref{fig:questionnaires}(right). 
Statistical analysis indicated that participants experienced significantly higher psycho-physical stress in condition \textit{a}, where the CoBot aimed solely at maximising productivity, compared to the other tested conditions ($\text{p}_{a,b}, \text{p}_{a,c} < 0.001$). 
In contrast, employing minimum-jerk robot movements (condition \textit{b}) led to the lowest stress level and can serve as a baseline reference for the stress experienced in HRC. Interestingly, the stress observed in condition \textit{c} was found to be comparable to that in condition \textit{b} ($\text{p}_{b,c} = 0.093$). 
The HRV analysis in the frequency domain further supported these findings. Different experimental conditions impacted the LF and HF parameters (expressed in Hz), with a predominant decrease and increase observed, respectively, throughout the tasks. The average levels of normalised LF/HF ratio for the minimum-time, minimum-jerk, and time-jerk-path optimised trajectories were 
$0.54$, $0.30$ and $0.36$, respectively. 

\subsubsection{Key Performance Indicators}
\autoref{tab:results_kpi} reports the outcome of the proposed Key Performance Indicators (KPIs). 
Based on the data collected in this study, when the CoBot tracked fixed-path minimum-time trajectories (condition \textit{a}), the human had a 27\% probability of making an error during the execution of each cycle. Conversely, when the CoBot executed fixed-path minimum-jerk motions (condition \textit{b}), the occurrence of errors during task completion was reduced, but the production rate severely decreased by 78\% ($\text{p}_{a,b} < 0.001$). In the tuned condition \textit{c}, the average number of errors was the same as in condition \textit{b}, but the production rate improved considerably ($\text{p}_{b,c} < 0.001$) but not to values comparable with condition \textit{a}. 

Interestingly, within the custom questionnaire (see the left side of \autoref{fig:questionnaires}), the participants rated the productivity of the trajectory adaptation strategy as optimal, similar to the condition where the CoBot solely focused on minimising time.

\begin{table}[!t]
\centering
\caption{Comparison of KPIs among testing conditions.}
\label{tab:results_kpi}
\begin{tabular}{@{}lll@{}}
\toprule
 & $\rho$ & $\epsilon$ \\
 & {\color[HTML]{676767} mean (std)} &
  {\color[HTML]{676767} mean (std)} \\
 \cmidrule(r){1-1} \cmidrule(r){2-3}
Minimum Time 
    & 2.17 
        {\hspace{0.78cm}\color[HTML]{676767}\rdelim]{1.4}{1mm} \multirow{2}{1.5mm}{**}} 
    & 0.27 (0.43) 
        {\color[HTML]{676767}\rdelim]{1.4}{1mm} \multirow{2}{1.5mm}{**}}
        {\color[HTML]{676767}\rdelim]{3}{1.5mm} \multirow{3}{1mm}{**}}\\
Minimum Jerk
    & 0.47  
    & 0.02 (0.07) \\
Tuned Online
    & 1.64 (0.37) 
        {\color[HTML]{676767}\rdelim]{-1.4}{1mm} \multirow{-1}{1.5mm}{**}} 
        {\color[HTML]{676767}\rdelim]{-3}{1.5mm} \multirow{-3}{1mm}{**}} 
    & 0.02 (0.06)   \\
    \bottomrule
\end{tabular}

\smallskip
Significance levels are indicated at the *$\text{p}<0.05$, **$\text{p}<0.001$.
\vspace{-0.3cm}
\end{table}

\section{Discussion}
\label{sec:discussion}

The results of the \textit{Path Adaptation Experiment} demonstrated the effectiveness of the proposed strategy in adapting the path of the CoBot during the execution to prevent it from violating subject-specific physical and cognitive-grounded safety. A strong point of these path alterations is their local nature, leveraging the properties of B-splines. 
This means that, without re-planning the entire trajectory, we can preserve path continuity and smoothness, which are crucial factors in promoting human comfort\cite{Flash1985}.  
Additionally, by using information on the operator's attention, mental effort and motion, the algorithm did not introduce any avoidable delays in the task accomplishment. As a result, our method allowed for faster task completion times and nearly eliminated robot idles compared to state-of-the-art techniques. 

An intriguing finding observed in \autoref{fig:continuity} (i.e. \textit{Overall Validation Experiment}) is the variation in individual needs during the execution of the activity, which can be influenced by cumulative fatigue, task demand, or even environmental factors (e.g. noise). To address this, our tuning algorithm online adjusts the pace of interaction based on the person-specific $\overline{\text{RR}}$ ranges and the estimated stress level experienced by the operator, in addition to varying safety distances based on human awareness. This allows for personalised CoBot trajectories, accommodating both inter- and intra-subject variations and thus surpassing the previous approaches \cite{kirschner2022expectable}.
The study outcomes support the hypothesis that relying solely on static timing law and fixed-distance safety zones is not optimal. In those conditions, CoBot can violate the comfort of na\"{i}ve or under-pressure users and prove overly cautious for individuals who are highly aware of their surroundings (e.g. by potentially leading to unnecessary protracted and lengthened CoBot routes or even halting its movement).  
Notably, the camera-based assessment of perilous work intensification anticipated stress detection and promoted the framework responsiveness to changes in the human psycho-physical state, reducing the overall stress experienced during the task compared to our previous study \cite{Lagomarsino2022robot}. 

The statistical analysis of HRV features and questionnaire responses revealed a significant reduction in stress and workload compared to the minimum-time trajectories condition. 
The experienced values were comparable to the minimum levels observed in HRC, specifically in the cosy condition where the CoBot followed minimum-jerk trajectories. 
However, the substantial velocity limitation of minimum-jerk motions resulted in human frustration, probably due to excessive caution even when it was unnecessary. This frustration was not observed with our trajectory adaptation strategy. 
This is a remarkable outcome indicating that the proposed framework is effective in mitigating psycho-physical stress in mixed human-robot environments while keeping the operator engaged in the task. 
Finally, the simultaneous path adaptation and time-jerk optimisation contributed to user comfort and fostered their perception of safety and performance. 

Regarding human-robot throughput, increasing the speed of the CoBot consistently results in a higher production rate. However, this affects the collaboration quality. Our strategy ensures excellent levels of average productivity without compromising human performance in terms of committed errors, imposed mental and physical demands, or induced stress.

The framework can be generalised to multiple industrial collaborative activities. It is noteworthy that manipulation tasks involve switching between constrained trajectories dictated by sub-task requirements (e.g. assuring zero end-effector velocity during object grasping), resulting in considerable variations of Cartesian velocity and acceleration. For this reason, it becomes challenging to force trajectory velocity and smoothness simultaneously, and thus we propose a multi-objective optimisation. Further strengths of the employed optimisation include eliminating the objectives' normalisation and initialisation procedure, which are crucial steps in single-objective problems with weighted terms.

\section{Conclusions}
\label{sec:conclusion}

This paper introduced PRO-MIND, a generalisable HRC strategy that enables CoBots to adjust their proximity level and reactive behaviour based on industrial-fit perception and monitoring of human attention and psycho-physical state.  
The integration of thorough data about human co-workers in the control loop distinguishes PRO-MIND as a novel and promising solution to rethink the safety limits and comfort in industrial human-robot collaboration. 
By online adapting safety zones around the operator and optimising the pace of interaction, PRO-MIND demonstrated significant advantages in enhancing individual cognitive resources and promptly responding to stress, thus maximising worker engagement and overall productivity of human-robot teams. 
The relevance and positive impact of the proposed framework are robustly supported by both subjective questionnaires and objective criteria, including physiological signal processing and common industrial KPIs.

In this study, the trajectory adaptation was constrained by the limited workspace of the employed fixed-base CoBot, and restrictions for safety requirements adherence remain.  
Future research can explore further enhancements of the framework including mobile CoBots and investigate its deployment in various unstructured environments to validate its generalisability and robustness. 
In addition, to boost its applicability in real industrial scenarios, the integration of additional safety-rated sensors, such as safety scanners, and the technical specifications outlined in \cite{iso15066} will be addressed. One potential approach could consider setting the radius of the physical safety zone equal to the robot braking distance as proposed in \cite{iso15066}, augmented by a parameter that accounts for human awareness. 
Moreover, extending the framework to ensure protective distancing from various human body parts and potential obstacles within the robot's workspace would be beneficial. 
These integrations would contribute to addressing the compliance of our framework with technical requirements while also aligning with the evolving needs of industrial safety. 

\balance
\bibliographystyle{IEEEtran}

\bibliography{bibliography}

\vfill

\end{document}